\documentclass[a4paper,fleqn]{cas-dc}
\usepackage{amssymb}
\usepackage{hyperref}
\usepackage{graphicx}
\usepackage[caption=false]{subfig}
\usepackage{lipsum}
\usepackage{lscape}
\usepackage{amsmath}
\usepackage{multirow}
\usepackage[figuresright]{rotating}
\usepackage{lineno}
\usepackage{algorithm} 
\usepackage{algpseudocode} 
\usepackage{comment}
\usepackage{lineno}
\usepackage{nomencl}
\makenomenclature
\usepackage{calc}
\usepackage[T1]{fontenc}
\usepackage{relsize}
\usepackage{subfig}
\usepackage{gensymb}
\usepackage{float}

\usepackage[sort,comma,authoryear,round]{natbib}

\setcitestyle{authoryear,open={(},close={)},citesep={;}} 

\hypersetup{
  colorlinks,
  citecolor=Violet,
  linkcolor=Red,
  urlcolor=Blue}

\newcolumntype{P}[1]{>{\centering\arraybackslash}p{#1}}

\def\tsc#1{\csdef{#1}{\textsc{\lowercase{#1}}\xspace}}
\tsc{WGM}
\tsc{QE}
\tsc{EP}
\tsc{PMS}
\tsc{BEC}
\tsc{DE}

\begin{document}
\let\WriteBookmarks\relax
\def\floatpagepagefraction{1}
\def\textpagefraction{.001}
\shorttitle{Foundation Models in Fruit Object Detection}
\shortauthors{Li et~al.}
 
\title [mode = title]{MetaFruit Meets Foundation Models: Leveraging a Comprehensive  Multi-Fruit Dataset for Advancing Agricultural Foundation Models}

\author[1]{Jiajia Li}\ead{lijiajia@msu.edu}
\author[1]{Kyle Lammers}\ead{lammer18@msu.edu}
\author[2]{Xunyuan Yin}\ead{xunyuan.yin@ntu.edu.sg}
\author[3]{Xiang Yin}\ead{yinxiang@sjtu.edu.cn}
\author[4]{Long He}\ead{luh378@psu.edu}
\author[5]{Renfu Lu}\ead{renfu.lu@usda.gov}
\author[6]{Zhaojian Li*}\ead{lizhaoj1@egr.msu.edu}

\address[1]{Department of Electrical and Computer Engineering, Michigan State University, East Lansing, MI, USA}
\address[2]{School of Chemical and Biomedical Engineering, Nanyang Technological University, Singapore}
\address[3]{Department of Automation and Key Laboratory of System Control and Information Processing, Shanghai Jiao Tong University, Shanghai, China}
\address[4]{Department of Agricultural and Biological Engineering, Pennsylvania State University, USA}
\address[5]{United States Department of Agriculture Agricultural Research Service, East Lansing, MI, USA}
\address[6]{Department of Mechanical Engineering, Michigan State University, East Lansing, MI, USA}

\address{* Zhaojian Li is the corresponding author}

\begin{abstract}
Fruit harvesting poses a significant labor and financial burden for the industry,
%
%
highlighting the critical need for advancements in robotic harvesting solutions. 
Machine vision-based fruit detection has been recognized as a crucial component for robust identification of fruits to guide robotic manipulation. Despite considerable progress in leveraging deep learning and machine learning techniques for fruit detection, a common shortfall is the inability to swiftly extend the developed models across different orchards and/or various fruit species. Additionally, the limited availability of pertinent data further compounds these challenges.
In this work, we introduce MetaFruit, the largest publicly available multi-class fruit dataset, comprising 4,248 images and 248,015 manually labeled instances across diverse U.S. orchards. 
Furthermore, this study proposes an innovative open-set fruit detection system leveraging advanced Vision Foundation Models (VFMs) for fruit detection that can adeptly identify a wide array of fruit types under varying orchard conditions.
This system not only demonstrates remarkable adaptability in learning from minimal data through few-shot learning but also shows the ability to interpret human instructions for subtle detection tasks. 
The performance of the developed foundation model is comprehensively evaluated using several metrics, which outperforms the existing state-of-the-art algorithms in both our MetaFruit dataset and other open-sourced fruit datasets, thereby setting a new benchmark in the field of agricultural technology and robotic harvesting. 
The MetaFruit dataset (\url{https://www.kaggle.com/datasets/jiajiali/metafruit}) and detection framework (\url{https://github.com/JiajiaLi04/FMFruit}) are open-sourced to foster future research in vision-based fruit harvesting, marking a significant stride toward addressing the urgent needs of the agricultural sector.
\end{abstract}


\maketitle

\section{Introduction}
\label{sec:intro}

\begin{table*}[!ht]
\renewcommand{\arraystretch}{1.6}
\centering
\caption{List of publicly available fruit datasets and our new MetaFruit dataset.  It details the data modality, total number of images (data numbers), count of instances per dataset, and the specific tasks each dataset supports. }
\label{tab:fruit_datasets}
\resizebox{0.98 \textwidth}{!}{%
\begin{tabular}{|l|l|c|c|c|c|c|l}
\hline
Datasets                                                                                                        & Fruit Variety                                                   & Modality & \# Images  & \# Instances           & Tasks                                   \\ \hline \hline
MangoNet \citep{kestur2019mangonet}                                                            & Mango                                                           & RGB      & 49      & -              & Fruit segmentation                      \\ 
MangoYOLO \citep{koirala2019deep}                                                              & Mango                                                           & RGB      & 1,730      & 9,067            & Fruit detection                  \\ 
DeepBlueberry \citep{gonzalez2019deepblueberry}                                                              & Blueberry                                                           & RGB      & 293      & 10,161            & Fruit detection                  \\ 
StrawDIDb1 \citep{perez2020fast}                                                              & Strawberry                                                           & RGB      & 3,100      & 17,938            & Fruit detection and segmentation                  \\ 
KFuji RGB-DS \citep{gene2019kfuji}                                                             & Apple                                                           & RGB-D    & 967    & 12,839                & Fruit detection                  \\ 

WSUApple \citep{bhusal2019apple}                                                               & Apple                                                           & RGB      & 2,298        & -           & Fruit detection                  \\ 
Fuji-SfM \citep{gene2020fuji}                                                            & Apple                                                           & RGB      &   288   &  1,455             & Fruit detection                  \\ 
LFuji-air dataset \citep{GENEMOLA2020105248}                                                        & Apple                                                           & LiDAR    & -     & -       & Fruit detection                  \\ 
MinneApple \citep{hani2020minneapple}                                                          & Apple                                                           & RGB      & 1,001      & 41,325            & Fruit detection and segmentation \\ 
OrchardFruit \citep{bargoti2017deep}                                                           & Apple, mango, and almond                                            & RGB      & 3,232      & -           & Fruit detection                  \\ 
DeepFruits \citep{sa2016deepfruits}                                                            & \begin{tabular}[c]{@{}l@{}} Strawberry, rockmelon, orange, mango, \\ capsicum, avocado, and apple \end{tabular} & RGB      & 587         & -         & Fruit detection                  \\ 
FruitNet \citep{meshram2022fruitnet}                                                           & \begin{tabular}[c]{@{}l@{}}  Apple, banana, guava, lime, orange, \\ and pomegranate \end{tabular}            & RGB      & \textgreater{}19,500 & -  & Fruit quality classification            \\ 
\begin{tabular}[c]{@{}l@{}}Fruit360 (\url{https://www.kaggle.com/datasets/moltean/fruits})\end{tabular} & 80 classes of fruits                                            & RGB      & 41,322  & -               & Fruit classification                    \\ \hline
MSUAppleDataset (Ours) \citep{chu2021deep}                                                                                                             &  Apple                     & RGB      & 1,500     & 19,528            & Fruit detection                  \\
MSUAppleDatasetv2 (Ours)  \citep{chu2023o2rnet}                                                                                                          &  Apple                     & RGB      &  1,246    & 14,518            & Fruit detection                  \\
MetaFruit (Ours)                                                                                                           &  Apple, orange, lemon, tangerine, grapefruit                     & RGB      & 4,248  & 248,015                 & Fruit detection                  \\ \hline
\end{tabular}
}
\end{table*}

Farm work is inherently labor-intensive and represents a significant burden. According to a report by the Economic Research Service of the U.S. Department of Agriculture\footnote{\url{https://www.ers.usda.gov/topics/farm-economy/farm-labor}}, farm labor in the U.S. relies heavily on immigrants, particularly those of Hispanic or Mexican origin. The H-2A Temporary Agricultural Program serves as a vital resource for crop farmers to address seasonal labor demands. Over the past 17 years, the number of H-2A positions requested and approved has increased more than sevenfold.  
Additionally, for all farms, labor costs averaged 10.4\% of gross cash income during 2018–2020, with the figure reaching about 30\% for fruits and tree nuts.
%
%
The mechanical harvesting system for fruits is an efficient and profitable approach but has the problem of excessive mechanical damage on fruits \citep{li2011review}. 
%
%
Therefore, there is a critical need for the innovation of robotic harvesting technologies to mitigate labor shortages, minimize human injury risks, and boost the efficiency and economic viability for the fruit industry \citep{sarig1993robotics, zhou2022intelligent}. 

The perception system is essential in harvesting robots, as it enables the identification of fruits within the target area and guides the robot in executing subsequent tasks \citep{zhao2016review, chu2021deep}. Recent significant advancements in the affordability of cameras and computer vision (CV) technology have made image-based fruit detection systems increasingly popular in robotic fruit harvesting \citep{gongal2015sensors, zhang2021system}. For instance, in \cite{syal2014apple}, the minimum Euclidean distance-based segmentation method is proposed for segmenting the fruit region from the input image to identify and count the number of fruits on trees. In \cite{chaivivatrakul2014texture}, conventional Machine Learning (ML)/CV techniques such as interest point feature extraction, support vector machines, and interest region extraction are developed for detecting green fruits on plants based on texture analysis. However, these methods depend heavily on manually designed features and can be negatively impacted by variations in lighting conditions and occlusions \citep{chu2021deep}.

More recently, deep learning (DL) based approaches have rapidly evolved and attracted significant attention in various agricultural sectors, such as plant disease identification \citep{xu2022style}, weed detection \citep{chen2024synthetic, rai2024weedvision, li2024performance}, plant counting \citep{li2024soybeannet}, and plant breeding \citep{li2024cotton}. These DL methods have also been proven effective in fruit detection \citep{koirala2019deep, ukwuoma2022recent, xiao2023fruit}. For instance, Faster-RCNN \citep{girshick2015fast} has been successfully applied for apple \citep{gao2020multi, fu2020faster}, kiwifruit \citep{fu2018kiwifruit}, and multiple fruits detection (mangoes, almonds and apples) \citep{bargoti2017deep}. 
In addition, YOLO models \citep{terven2023comprehensive} are also applied for fruit detection and recognition such as apple \citep{tian2019apple}, mango \citep{shi2020attribution}, orange \citep{mirhaji2021fruit}, and cherry \citep{gai2023detection}. 
In our previous research, state-of-the-art DL techniques based on Mask-RCNN \citep{he2017mask} and Faster RCNN \citep{girshick2015fast} are developed for accurate apple detection for dense orchard settings \citep{chu2021deep, chu2023o2rnet}. 
Despite the aforementioned successes, developing DL models from scratch faces several challenges. Firstly, it relies heavily on large, accurately annotated image datasets, which are generally costly to obtain \citep{li2023label, li2023ml}. Secondly, the training phase is remarkably time-intensive and demands significant computational resources \citep{lecun2015deep}.  Moreover, while these specialized models excel in their designated tasks, they often encounter difficulties when applied to novel scenarios, such as different orchard conditions or fruit species, demonstrating limited capabilities in generalization \citep{kamilaris2018deep}.

It is widely acknowledged that a comprehensive set of annotated images is essential for the development of high-performing DL models in visual fruit detection tasks \citep{sun2017revisiting}. In \cite{lu2020survey}, the authors have provide an overview of various publicly accessible fruit image datasets aimed at robotic harvesting. 
For instance, mango-related datasets, such as MangoNet \citep{kestur2019mangonet} and MangoYOLO \citep{koirala2019deep} contain 49 and 1730 images for mango segmentation and detection, respectively.  There are specialized apple datasets for apple detection, including KFuji RGB-DS \citep{gene2019kfuji} WSUApple \citep{bhusal2019apple}, LFuji-air dataset \citep{gene2020lfuji}, and MinneApple \citep{hani2020minneapple}, along with two apple datasets from our previous studies \citep{chu2021deep, chu2023o2rnet}. Additionally, DeepBlueberry \citep{gonzalez2019deepblueberry} is a dataset including 294 images for blueberry detection. However, most of these datasets are species-specific and not transferable to different fruit types. 
Recently, there has been an increasing interest in multi-fruit datasets. For instance, FruitNet \citep{meshram2022fruitnet} and Fruit360\footnote{\url{https://www.kaggle.com/datasets/moltean/fruits}} feature 19,500 and 41,322 images across 5 and 80 fruit species, respectively, catering to fruit classification tasks. In terms of fruit detection, OrchardFruit \citep{bargoti2017deep} and DeepFruits \citep{sa2016deepfruits} provide open-source access to 3,232 and 587 images for 3 and 7 fruit species, respectively. Yet, these datasets are typically designed for specific orchard environments with less dense fruit clusters. Table~\ref{tab:fruit_datasets} summarizes these datasets, providing an overview of the resources available for advancing research in fruit detection.

Lately, the rise of large pre-trained models, commonly known as foundation models (FMs), such as ChatGPT-4 \citep{achiam2023gpt}, Segment Anything Model (SAM) \citep{kirillov2023segment}, have demonstrated outstanding performance in both language and vision tasks across diverse domains \citep{bommasani2021opportunities, li2023foundation}. These models undergo extensive training on diverse datasets spanning multiple domains and modalities. Once fully trained, they exhibit the capability to perform a range of tasks requiring minimal fine-tuning and without extensive reliance on task-specific labeled data. There has been growing interest in applying FMs within the field of agriculture, offering innovative solutions and insights. 
As an example, \cite{yang2023sam} employs SAM for chicken segmentation tasks in a zero-shot manner, integrating part-based segmentation and the use of infrared thermal imagery. 
The experimental findings reveal that SAM outperforms other vision foundation models (VFMs) like SegFormer and SETR in accuracy for both whole and partial chicken segmentation. 
\cite{williams2023leaf} introduce ``Leaf Only SAM'', an automatic leaf segmentation pipeline designed for zero-shot segmentation of potato leaves. Compared to a fine-tuned Mask R-CNN model tailored for annotated potato leaf datasets, this innovative approach demonstrates superior effectiveness. 
%
These developments underscore the potential of FMs in various agricultural applications. However, to the best of our knowledge, FMs have not yet been applied to fruit harvesting tasks involving multiple fruit classes.

In this study, we introduce a comprehensive multi-class fruit dataset (also named MetaFruit), gathered from commercial orchards in two U.S. states with greatly different geographic locations during the growth seasons of 2022 and 2023. Building on this, we develop an innovative open-set fruit detection system, leveraging the power of advanced vision FMs (VFMs) to identify a wide range of fruits. The contributions of this work can be summarized as follows:
\begin{enumerate}
    \item We introduce a very comprehensive and diverse fruit dataset, including 4,248 images with 248,015 manually labeled fruit instances, meticulously collected from commercial orchard fields across two U.S. states.
    \item We propose a novel FM-based open-set fruit detection framework designed for multi-class fruit detection, which is not only capable of identifying various and novel types of fruit but also integrates the ability to process human language inputs.
    \item Comprehensive experiments are conducted to rigorously assess the performance of our proposed framework, performing not only on our newly collected dataset but also on existing open-sourced fruit datasets.
    \item Both curated dataset\footnote{\url{https://www.kaggle.
com/datasets/jiajiali/metafruit}} and developed software\footnote{\url{https://github.com/JiajiaLi04/FMFruit}} are open-sourced, making them accessible for further research and engineering integration in vision-based fruit harvesting and related applications.
\end{enumerate}

\begin{figure*}[!ht]
  \centering
  \includegraphics[width=0.95\textwidth]{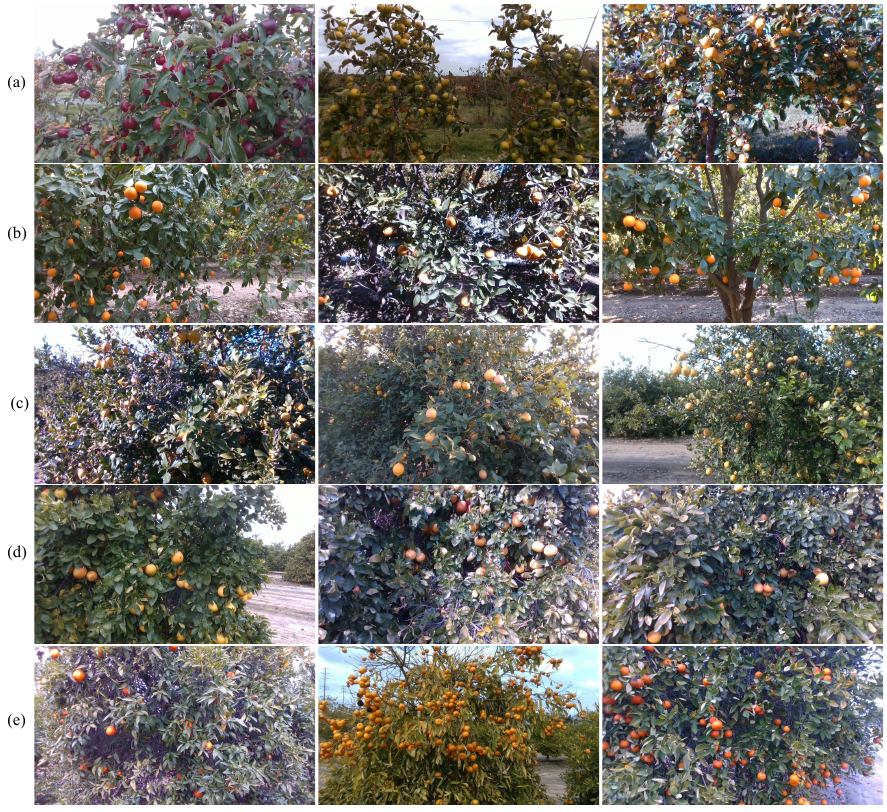}
  \caption{Representative examples of MetaFruit dataset, including five fruit classes: (a) apple, (b) orange, (c) lemon, (d) grapefruit, and (e) tangerine.}
  \label{fig:metafruit}
\end{figure*}

\section{Materials and Methods}
\label{sec:matrials}
In this section, we first present our collected dataset, MetaFruit, and the VFMs used for multi-class fruit detection. We then detail the few-shot learning, evaluation metrics, and experimental setups employed in our study.

\subsection{MetaFruit dataset}
The multi-class fruit dataset, MetaFruit, introduced in this study is collected utilizing advanced imaging technology, comprising both a high-definition camera and a sophisticated LiDAR system (with a resolution of $1920 \times 1080$), from commercial orchards in North Michigan and California, USA. To guarantee a diverse and varied collection of images that enhances model robustness \citep{lu2020survey}, the dataset includes images taken under natural field lighting conditions across various weather conditions (e.g., sunny, cloudy, and overcast) during the peak harvest season of the fruit growth stage. The dataset contains 4,247 images, featuring five distinct fruit types: apples, oranges, lemons, grapefruits, and tangerines. Figure~\ref{fig:metafruit} shows representative samples for each fruit category. Unlike existing datasets, MetaFruit is characterized by more realistic/complex orchard environments with fruits frequently appearing in clusters, presenting a challenging yet realistic scenario for model training and evaluation. Notably, the dataset also includes multiple varieties within each fruit category. For example, the apple class includes both red and green species, adding another layer of diversity and complexity to the dataset. 

The images acquired for the MetaFruit dataset are meticulously labeled by trained personnel. These annotators utilized the Labelme \citep{Wada_Labelme_Image_Polygonal} tool to accurately draw bounding boxes around individual fruit instances in the images. This meticulous process results in the acquisition of 248,015 manually labeled bounding boxes. The distribution of the MetaFruit dataset is detailed in Table~\ref{tab:dataset_intro}. Overall, the dataset exhibits an even distribution among apples, oranges, lemons, and tangerines, each with a similar number of images, whereas grapefruits are represented with slightly fewer images, totaling 490. Tangerines are particularly well-represented in the dataset with 1,063 images and 85,785 labeled instances, averaging 81 bounding boxes per image. The average number of bounding boxes per image sheds light on the density of fruits captured in the images, whereas the average size of these instances provides insight into the physical size of the objects. Notably, the smaller the size of the instances, the greater the challenge in detecting them accurately. Interestingly, while the lemon class does not have the highest average number of bounding boxes per image, it features the smallest average size of instances (823 pixels per instance), indicating lemons' smaller physical presence within the images, which presents its unique detection challenges.

The MetaFruit dataset, in terms of instance numbers, significantly surpasses previous collections, being more than 10 times larger than the dataset for multi-class fruit species featured in OrchardFruit \citep{bargoti2017deep} (as shown in Table~\ref{tab:fruit_datasets}). To the best of our knowledge, it represents the most extensive publicly available dataset for fruit detection specifically designed for commercial orchard systems, establishing a new benchmark for research and development in agricultural technology and robotic harvesting.

\begin{table*}[!ht]
\renewcommand{\arraystretch}{1.4}
\centering
\caption{Statistics of MetaFruit dataset. This includes the total number of images, the total number of bounding boxes, the average number of bounding boxes per image, the average size of each instance (measured in pixels), and the geographical region of data collection for each fruit type.}
\label{tab:dataset_intro}
\resizebox{0.74 \textwidth}{!}{%
\begin{tabular}{|l|c|c|c|c|c|}
\hline
\multicolumn{1}{|l|}{} & \# imgs & \# bboxes & \# avg. bboxes/image & \# avg. size/instance & Region \\ \hline \hline
Apple                  & 812     & 62,040 & 76 & 1,193 &  Michigan \& California   \\ 
Orange                 & 926     & 45,834  & 49 & 1,178 & California  \\ 
Lemon                  & 958     & 42,238 & 44  & 823 & California  \\ 
Grapefruit             & 490     & 12,118 & 25 & 2,232 & California   \\ 
Tangerine              & 1,062    & 85,785 & 81 & 1,068 & California   \\ \hline
Total                  & 4,248    & 248,015 & 58 & 1,133 & Michigan \& California  \\ \hline
\end{tabular}
}
\end{table*}

\begin{figure*}[!ht]
  \centering
  \includegraphics[width=0.9\textwidth]{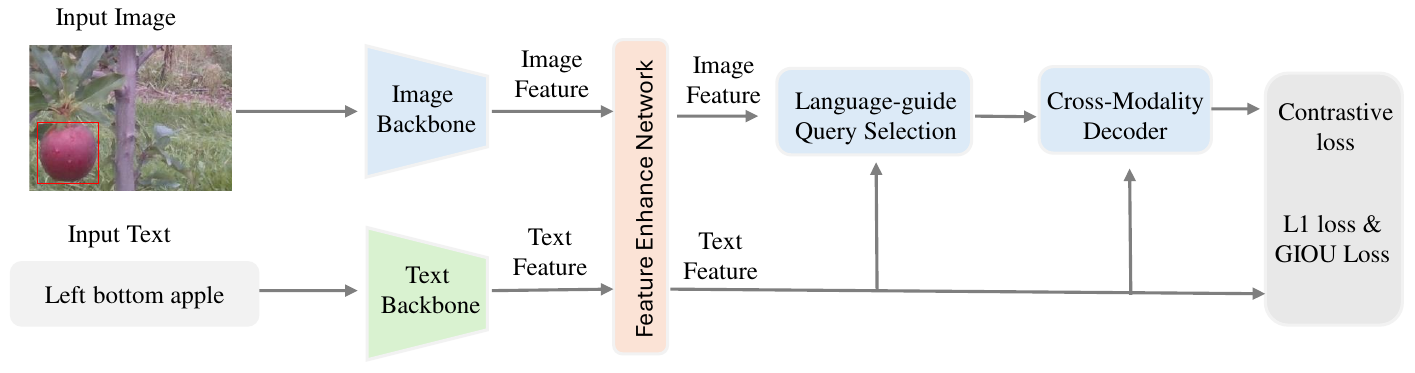}
  \caption{The framework of the VFM for fruit detection based on the Grounding DINO \citep{liu2023grounding} model.}
  \label{fig:dino}
\end{figure*}

\subsection{VFMs for fruit detection}
In recent years, DL approaches have made significant strides in advancing fruit detection models. Prominent among the object detectors employed are FCOS \citep{tian2020fcos}, Faster-RCNN \citep{girshick2015fast}, and YOLO series \citep{terven2023comprehensive}, all of which are designed as closed-set detectors. Such models operate under the assumption that the categories of objects to be detected are predefined and known during both the training and testing phases, thereby limiting their capacity to recognize previously unseen categories. Furthermore, these approaches depend on extensive, meticulously labeled image datasets—a process that is both labor-intensive and demands significant resources. 
In contrast, recent focus has shifted towards open-set object detection  \citep{geng2020recent} and the exploration of LLMs and FMs \citep{bommasani2021opportunities, li2023ml, li2023foundation}. These open-set detectors are capable of not only precisely detecting the known classes but also efficiently handling the unknown ones.
Therefore, the language data needs to be added for model training to solve the situation that a testing sample comes from some unknown classes. 
Similarly, LLMs and FMs, which are trained on extensive datasets covering a wide range of domains and modalities, demonstrate a remarkable ability to perform a variety of open-set tasks after training, which is achieved with minimal fine-tuning and reduced reliance on extensive, task-specific labeled data. 

To facilitate open-set fruit detection across a diverse array of fruit categories, this study employs a vision foundation model (VFM), specifically the Grounding DINO \citep{liu2023grounding} model, for the detection task. Grounding DINO is an open-set detector predicated on the DETR-like architecture, DINO \citep{zhang2022dino}, which integrates end-to-end Transformer-based detection mechanisms. A pivotal aspect of enabling open-set detection capabilities is the integration of linguistic elements for the generalization of unseen objects. This approach involves training the model on existing bounding box annotations, augmented through language generalization, to facilitate the identification of a broader array of objects beyond those seen during training. 

The overall workflow and architectural design are illustrated in Figure~\ref{fig:dino}. Initially, the process involves extracting fundamental features from both images and text through respective image and text backbones, i.e., the Swin Transformer \citep{liu2021swin} module. These foundational features serve as inputs to a feature enhancer network dedicated to the fusion of cross-modality features, facilitating a comprehensive integration of image and textual information. Following the acquisition of enriched cross-modality text and image features, the system employs a language-guided query selection module \citep{liu2023grounding} to meticulously select cross-modality queries based on the image features, thereby harnessing the synergistic potential of linguistic cues and visual data. This selection process mirrors the transformative approach of integrating diverse modalities to enhance detection precision and contextual understanding through the strategic alignment of textual and visual elements. Subsequently, these cross-modal queries are introduced into a cross-modal decoder to extract and refine the desired features from the combined bimodal information, continually updating its parameters to reflect the insights gained from the cross-modal analysis. Ultimately, the decoder's output queries are used to predict object bounding boxes and identify relevant textual phrases, culminating in a sophisticated system capable of precise object detection and association with appropriate linguistic descriptors. 
The loss function is defined as 
\begin{equation}
    \mathcal{L} = \mathcal{L}_1 + \mathcal{L}_{\text{GIOU}} + \mathcal{L}_{\text{Cons}},
\end{equation}
where $\mathcal{L}_1$ and $\mathcal{L}_{\text{GIOU}}$ \citep{rezatofighi2019generalized} are utilized for the regression of bounding boxes. The contrastive loss, $\mathcal{L}_{\text{Cons}}$, is also incorporated as  in GLIP \citep{li2022grounded}, to fine-tune the classification of predicted objects and language tokens \citep{liu2023grounding, zhang2022dino}.

The Grounding DINO model \citep{liu2023grounding} leverages the foundational DINO architecture \citep{zhang2022dino}, and to save computational resources and training time, the Grounding DINO model is transferred from DINO weights instead of training from scratch \citep{zhuang2020comprehensive}. The DINO model is trained on the O365 data \citep{shao2019objects365}, which is a large-scale object detection dataset containing 365 categories and 2 million images. 
Based on the pre-trained DINO weights, the grounding DINO with swin-transformer tiny backbone is trained on a combined data set including O365, GoldG, and Cap4M, where GoldG contains images in Flickr30k entities \citep{plummer2015flickr30k} and Visual Genome \citep{krishna2017visual}, and Cap4M is from \citep{li2022grounded} but not publicly available. 
Similarly, the grounding DINO with Swin-transformer large backbone is also transferred from DINO, but with more data (e.g., O365, GoldG, Cap4M, OI \citep{krasin2017openimages}, RefCOCO/+/g \citep{kazemzadeh2014referitgame}, and COCO).  
To tailor the Grounding DINO model for the specific task of detecting a wide array of fruits in open-set conditions, we conduct fine-tuning using our MetaFruit dataset based on the pre-trained Grounding DINO weights, which is referred to FMFruit in the following sections.

\subsection{Few-shot learning}
Contemporary fruit detection algorithms, while yielding promising results, often struggle to generalize across varying data distributions, such as different fruit classes and orchard settings, especially when faced with a lack of extensive data \citep{wang2020generalizing}. The scarcity of data can be attributed not only to the inherent challenges of the task or privacy issues but also to the significant costs associated with data preparation, including collection, preprocessing, and labeling. In response to these challenges, few-shot learning has gained recognition as a promising learning method, demonstrating the significant potential for quickly learning underlying patterns from merely a few or even zero samples \citep{song2023comprehensive}. Zero-shot transfer learning refers to scenarios where no training samples are utilized, and models are directly deployed on testing images, aiming to make accurate predictions based solely on their pre-existing knowledge and capabilities. On the other hand, few-shot learning involves using a minimal number of samples to refine and adjust the models. For example, in 5-shot learning, precisely five samples are employed for model fine-tuning. It is important to note that while few-shot learning allows models to adapt to new tasks with limited data, the performance of such models, when only a few samples are used for fine-tuning, can sometimes be constrained. The effectiveness of the fine-tuning process is heavily dependent on the quality and representativeness of the selected samples, their alignment with the task at hand, and the model's inherent ability to generalize from minimal information \citep{wang2020generalizing, song2023comprehensive}. This delicate balance between sample selection and model adaptability is critical for maximizing the potential of few-shot learning approaches in diverse application scenarios, including those within the domain of fruit detection where variability across classes and environments is high.

In this study, we employ few-shot learning frameworks to evaluate the generalizability of the FMFruit model across various fruit categories. Specifically, the zero-shot learning scenario is utilized by deploying the FMFruit model on new fruit classes without any model fine-tuning. Concurrently, for the few-shot learning experiments, a minimal number of samples are randomly selected from these new fruit categories to slightly adjust the model.

\subsection{Evaluation metrics}
The performance of DL models in fruit detection tasks is rigorously evaluated using key detection accuracy metrics, such as Average Precision (AP), mean Average Recall (mAR), and mean Average Precision (mAP) \citep{dang2023yoloweeds}. These metrics collectively offer a detailed assessment of a model's proficiency in both identifying and precisely locating fruits within images. AP, with a specific focus on precision at a 50\% overlap threshold (AP50), and mAP, which calculates the average precision across a range of overlap thresholds (from 0.5 to 0.95, in increments of 0.05), together provide insights into the precision aspects of model performance. Meanwhile, mAR evaluates the model's recall capabilities over a spectrum of Intersection over Union (IoU) ranging from 0.5 to 0.95, thereby gauging the model's effectiveness in capturing the true positive detections across various conditions.

\subsection{Experimental setups}
Extensive experiments are conducted based on the following four settings: 
\begin{itemize}
    \item Zero-shot transfer, few-shot learning, and fine-tuning on our MetaFruits.
    \item Cross-class generalization ability evaluation by fine-tuning with four kinds of fruits and evaluating on the remaining novel one. 
    \item Zero-shot evaluation, few-shot learning, and fine-tuning on some of the existing fruit data.
    \item Case study of language-referring object detection.
\end{itemize}
We have two model variants, FMFruit-T with Swin-T \citep{liu2021swin}, and FMFruit-L with Swin-L \citep{liu2021swin} as the image backbone, respectively. Following BERT-base \citep{devlin2018bert}, Hugging Face \citep{wolf2019huggingface} is used as the text backbone. 
All the models are trained for 100 epochs with the AdamW optimizer. The learning rate is set to be 1e-4 with the weight decay as 0.0001, but the learning rate for the image and text backbone is set to be 1e-5. To expedite the model training process, we leverage transfer learning based on pre-trained DINO and pre-trained Grounding DINO \citep{zhuang2020comprehensive}. The fine-tuning procedure involves using a batch size of 4 over 100 epochs, and we utilize the PyTorch framework (version 1.10.1) \citep{paszke2019pytorch}. The MetaFruit dataset is divided into training and test sets, with a distribution ratio of 60\% for training and 40\% for testing. Both the training and testing phases of the models take place on a server running Ubuntu 20.04. This server is equipped with two GeForce RTX 2080Ti GPUs, each offering 12GB of GDDR6X memory.

\begin{table*}[!ht]
\renewcommand{\arraystretch}{1.4}
\centering
\caption{Zero-shot and few-shot performance on our MetaFruit dataset. All models are trained on our MetaFruit data. In the fine-tuning setting, the entirety of the available training data is utilized. Conversely, zero-shot learning does not involve a training process, relying instead on pre-existing model knowledge. For few-shot learning, only a few samples are used as training data. For instance, in the case of 1-shot learning for apples, training is conducted using just a single image. }
\label{tab:few_shot}
\resizebox{0.89 \textwidth}{!}{%
\begin{tabular}{|cc|ccc|ccc|ccc|ccc|ccc|}
\hline
\multicolumn{2}{|c|}{}                                                                & \multicolumn{3}{c|}{Apple}                                      & \multicolumn{3}{c|}{Orange}                                                             & \multicolumn{3}{c|}{Lemon}                                      & \multicolumn{3}{c|}{Grapefruit}                                 & \multicolumn{3}{c|}{Tangerine}                             \\  
\multicolumn{2}{|c|}{\multirow{-2}{*}{}}                                              & \multicolumn{1}{c}{mAP}   & \multicolumn{1}{c}{AP50}  & mAR   & \multicolumn{1}{c}{mAP}   & \multicolumn{1}{c}{AP50}                          & mAR   & \multicolumn{1}{c}{mAP}   & \multicolumn{1}{c}{AP50}  & mAR   & \multicolumn{1}{c}{mAP}   & \multicolumn{1}{c}{AP50}  & mAR   & \multicolumn{1}{c}{mAP}   & \multicolumn{1}{c}{AP50}  & mAR  \\ \hline \hline
\multicolumn{1}{|c|}{Retinanet}                         &                             & \multicolumn{1}{c}{38.0} & \multicolumn{1}{c}{67.4} & 42.9 & \multicolumn{1}{c}{41.2} & \multicolumn{1}{c}{\cellcolor[HTML]{FFFFFF}70.6} & 46.0 & \multicolumn{1}{c}{37.8} & \multicolumn{1}{c}{68.3} & 43.6 & \multicolumn{1}{c}{43.8} & \multicolumn{1}{c}{81.3} & 51.0 & \multicolumn{1}{c}{37.1} & \multicolumn{1}{c}{62.1} & 40.6    \\  
\multicolumn{1}{|c|}{Faster-RCNN}                       &                             & \multicolumn{1}{c}{48.4} & \multicolumn{1}{c}{78.4} & 53.3 & \multicolumn{1}{c}{50.9} & \multicolumn{1}{c}{83.1}                         & 55.6 & \multicolumn{1}{c}{46.7} & \multicolumn{1}{c}{78.6} & 52.7 & \multicolumn{1}{c}{51.3} & \multicolumn{1}{c}{86.8} & 56.5 & \multicolumn{1}{c}{43.9} & \multicolumn{1}{c}{70.3} & 47.4  \\  
\multicolumn{1}{|c|}{FCOS}                              &                             & \multicolumn{1}{c}{49.8} & \multicolumn{1}{c}{80.9} & 55.3 & \multicolumn{1}{c}{52.5} & \multicolumn{1}{c}{85.1}                         & 58.4 & \multicolumn{1}{c}{47.5} & \multicolumn{1}{c}{80.1} & 54.2 & \multicolumn{1}{c}{54.5} & \multicolumn{1}{c}{90.3} & 61.5 & \multicolumn{1}{c}{45.8} & \multicolumn{1}{c}{71.4} & 49.7  \\  
\multicolumn{1}{|c|}{RTMDet}                            & \multirow{-4}{*}{Fine-tuning} & \multicolumn{1}{c}{52.4} & \multicolumn{1}{c}{81.5} & 58.0 & \multicolumn{1}{c}{53.5} & \multicolumn{1}{c}{83.5}                         & 59.4 & \multicolumn{1}{c}{49.0} & \multicolumn{1}{c}{79.8} & 55.7 & \multicolumn{1}{c}{60.3} & \multicolumn{1}{c}{91.2} & 66.9 & \multicolumn{1}{c}{46.9} & \multicolumn{1}{c}{71.0} & 50.4  \\ \hline
\multicolumn{1}{|c|}{}                                  & Zero-shot                   & \multicolumn{1}{c}{24.1} & \multicolumn{1}{c}{45.7} & 46.1 & \multicolumn{1}{c}{36.6} & \multicolumn{1}{c}{68.9}                         & 52.2 & \multicolumn{1}{c}{29.4} & \multicolumn{1}{c}{52.1} & 47.3 & \multicolumn{1}{c}{37.2} & \multicolumn{1}{c}{64.4} & 52.7 & \multicolumn{1}{c}{29.6} & \multicolumn{1}{c}{60.3} & 43.8   \\  
\multicolumn{1}{|c|}{}                                  & 1-shot                      & \multicolumn{1}{c}{45.5} & \multicolumn{1}{c}{81.8} & 55.5 & \multicolumn{1}{c}{45.9} & \multicolumn{1}{c}{81.3}                         & 54.7 & \multicolumn{1}{c}{37.5} & \multicolumn{1}{c}{72.3} & 49.3 & \multicolumn{1}{c}{48.6} & \multicolumn{1}{c}{83.1} & 59.7 & \multicolumn{1}{c}{42.0} & \multicolumn{1}{c}{83.6} & 47.3   \\  
\multicolumn{1}{|c|}{}                                  & 5-shot                      & \multicolumn{1}{c}{48.0} & \multicolumn{1}{c}{85.2} & 56.5 & \multicolumn{1}{c}{48.4} & \multicolumn{1}{c}{82.7}                         & 56.8 & \multicolumn{1}{c}{42.7} & \multicolumn{1}{c}{76.3} & 52.6 & \multicolumn{1}{c}{47.0} & \multicolumn{1}{c}{81.9} & 58.8 & \multicolumn{1}{c}{38.6} & \multicolumn{1}{c}{79.2} & 46.2   \\  
\multicolumn{1}{|c|}{}                                  & 10-shot                     & \multicolumn{1}{c}{53.2} & \multicolumn{1}{c}{89.8} & 59.3 & \multicolumn{1}{c}{52.4} & \multicolumn{1}{c}{85.9}                         & 59.9 & \multicolumn{1}{c}{48.4} & \multicolumn{1}{c}{81.2} & 56.0 & \multicolumn{1}{c}{54.6} & \multicolumn{1}{c}{88.8} & 62.9 & \multicolumn{1}{c}{44.9} & \multicolumn{1}{c}{87.8} & 49.3  \\  
\multicolumn{1}{|c|}{}                                  & 20-shot                     & \multicolumn{1}{c}{55.3} & \multicolumn{1}{c}{91.3} & 60.9 & \multicolumn{1}{c}{54.4} & \multicolumn{1}{c}{87.4}                         & 61.7 & \multicolumn{1}{c}{49.8} & \multicolumn{1}{c}{82.9} & 57.4 & \multicolumn{1}{c}{57.8} & \multicolumn{1}{c}{91.0} & 65.6 & \multicolumn{1}{c}{46.7} & \multicolumn{1}{c}{90.3} & 50.9  \\  
\multicolumn{1}{|c|}{\multirow{-6}{*}{FMFruit-T}} & Fine-tuning                   & \multicolumn{1}{c}{\textbf{59.4}} & \multicolumn{1}{c}{\textbf{94.1}} & \textbf{64.7} & \multicolumn{1}{c}{\textbf{60.1}} & \multicolumn{1}{c}{\textbf{92.0}}                         & \textbf{66.5} & \multicolumn{1}{c}{\textbf{56.0}} & \multicolumn{1}{c}{\textbf{88.0}} & \textbf{62.6} & \multicolumn{1}{c}{\textbf{64.0}} & \multicolumn{1}{c}{\textbf{94.7}} & \textbf{70.9} & \multicolumn{1}{c}{\textbf{50.4}} & \multicolumn{1}{c}{\textbf{93.7}} & \textbf{54.4}  \\ \hline
\end{tabular}
}
\end{table*}

\section{Results}
In this section, we first evaluate the zero-shot and few-shot transfer learning performance of FMFruit in comparison with leading fruit detection algorithms on our MetaFruit dataset. Then, we examine its ability of cross-class generalization and evaluate its effectiveness on other publicly available fruit datasets. Lastly, we present initial findings on its capability to integrate text inputs and comprehend referring expressions.

\subsection{Few-shot fruit detection performance}
In this subsection, we examine the zero-shot and few-shot transfer learning capabilities of our proposed model across five distinct fruit types from our MetaFruits data. We compare our model's performance with that of leading object detection models, including Fully Convolutional One-Stage (FCOS) object detector \citep{tian2020fcos},  Faster-RCNN \citep{girshick2015fast}, RetinaNet \citep{lin2017focal}, and RTMDet \citep{lyu2022rtmdet}. The performance comparison is presented in Table~\ref{tab:few_shot}.  It is noteworthy that traditional CNN-based models such as FCOS, despite being trained on the comprehensive COCO dataset \citep{lin2014microsoft}, which encompasses 80 categories including apples and oranges, fail to achieve any positive mAP and mAR scores in fruit detection tasks across all fruit classes. This highlights a critical limitation of conventional object detection algorithms, which struggle with generalization across diverse datasets and are typically fine-tuned for narrow, specific detection scenarios. Among the baseline models, RTMDet emerges as one of the best-performing models following comprehensive training across all evaluated fruit types in terms of mAP and mAR metrics, while RetinaNet is observed to lag behind the rest of the baseline models in performance.

Conversely, our foundation model-based fruit detection model, FMFruit, demonstrates exceptional zero-shot transfer performance across all evaluated fruit classes. Notably, for FMFruit-T, two out of the five fruit classes achieve a mAP score exceeding 36.0, alongside an AP50  score and a mAR surpassing 64.0 and 52.0 across all types of fruits, respectively. The zero-shot experimental results underline its impressive capability to accurately detect and identify a wide range of fruits without specific prior training in those classes.
FMFruit-T's performance on apples shows a specific challenge, achieving a zero-shot 24.1 mAP score. This performance can be attributed to the presence of densely clustered fruits, with an average of 76 apples per image, as detailed in Table~\ref{tab:dataset_intro}. Similarly, the model's detection capability for lemons, which achieves a 29.4 mAP score, highlights the difficulty in accurately identifying fruits that occupy small areas within images, with the average size being only 823 pixels per lemon, as also indicated in Table~\ref{tab:dataset_intro}.

In few-shot learning scenarios, FMFruit-T exhibits promising performance across all fruit classes. In the 1-shot setting, four out of the five fruit classes achieve mAP, AP50, and mAR exceeding 42.0, 81.0, and 47.0, respectively, using just a single image per fruit class for training. Expanding to a 5-shot scenario, where five images per class are used for training, FMFruit-T maintains excellent performance, akin to the fine-tuning setting where all available images are used for training. Specifically, FMFruit-T yields significant improvement in the apple class performance, with AP50 increasing from 45.7 to 81.8 under the 1-shot setting compared with zero-shot learning. Moreover, with the 10-shot setting, it achieves a significant performance improvement, with an AP50 of 89.9, further illustrating the model's impressive ability to rapidly adapt and excel with minimal training data.

Additionally, the performance is compared between different backbones and pre-trained data for our FMFruit model. FMFruit-L is pre-trained with more data and a large backbone, which results in a better zero-shot and 1-shot performance. As shown in Table~\ref{tab: different_backbone}, FMFruit-L archives a mAP greater than 30 for all five types of fruits, contrasting with FMFruit-T, which achieves this threshold only for two fruit categories under the zero-shot learning. Furthermore, FMFruit-L outperforms FMFruit-T in 1-shot learning, whereas their performance is comparable in fine-tuning learning scenarios. In the fine-tuning setting, FMFruit-L performs slightly worse than FMFruit-T. Due to its model complexity and computational demands, FMFruit-L is more prone to suboptimal performance.

\begin{table*}[!ht]
\renewcommand{\arraystretch}{1.4}
\centering
\caption{Evaluation of different backbones and pre-trained data for our FMFruit model. The FMFruit-T model utilizes Swin-T as its backbone, while the FMFruit-L model employs Swin-L as the backbone. Additionally, FMFruit-L leverages more dataset for pre-training, enhancing its learning capabilities. }
\label{tab: different_backbone}
\resizebox{0.85 \textwidth}{!}{%
\begin{tabular}{|cc|ccc|ccc|ccc|ccc|ccc|}
\hline
\multicolumn{2}{|c|}{}                                         & \multicolumn{3}{c|}{Apple}                                      & \multicolumn{3}{c|}{Orange}                                                             & \multicolumn{3}{c|}{Lemon}                                      & \multicolumn{3}{c|}{Grapefruit}                                 & \multicolumn{3}{c|}{Tangerine}                                  \\ 
\multicolumn{2}{|c|}{\multirow{-2}{*}{}}                       & \multicolumn{1}{c}{mAP}   & \multicolumn{1}{c}{AP50}  & mAR   & \multicolumn{1}{c}{mAP}   & \multicolumn{1}{c}{AP50}                          & mAR   & \multicolumn{1}{c}{mAP}   & \multicolumn{1}{c}{AP50}  & mAR   & \multicolumn{1}{c}{mAP}   & \multicolumn{1}{c}{AP50}  & mAR   & \multicolumn{1}{c}{mAP}   & \multicolumn{1}{c}{AP50}  & mAR   \\ \hline \hline
\multicolumn{1}{|c|}{}                             & Zero-shot & \multicolumn{1}{c}{24.1} & \multicolumn{1}{c}{45.7} & 46.1 & \multicolumn{1}{c}{36.6} & \multicolumn{1}{c}{\cellcolor[HTML]{FFFFFF}68.9} & 52.2 & \multicolumn{1}{c}{29.4} & \multicolumn{1}{c}{52.1} & 47.3 & \multicolumn{1}{c}{37.2} & \multicolumn{1}{c}{64.4} & 52.7 & \multicolumn{1}{c}{29.6} & \multicolumn{1}{c}{60.3} & 43.8 \\  
\multicolumn{1}{|c|}{}                             & 1-shot    & \multicolumn{1}{c}{45.5} & \multicolumn{1}{c}{81.8} & 55.5 & \multicolumn{1}{c}{45.9} & \multicolumn{1}{c}{81.3}                         & 54.7 & \multicolumn{1}{c}{37.5} & \multicolumn{1}{c}{72.3} & 49.3 & \multicolumn{1}{c}{48.6} & \multicolumn{1}{c}{83.1} & 59.7 & \multicolumn{1}{c}{42.0} & \multicolumn{1}{c}{83.6} & 47.3 \\  
\multicolumn{1}{|c|}{\multirow{-3}{*}{FMFruit-T}} & Fine-tuning & \multicolumn{1}{c}{\textbf{59.4}} & \multicolumn{1}{c}{\textbf{94.1}} & \textbf{64.7} & \multicolumn{1}{c}{\textbf{60.1}} & \multicolumn{1}{c}{\textbf{92.0}}                         & 66.5 & \multicolumn{1}{c}{\textbf{56.0}} & \multicolumn{1}{c}{\textbf{88.0}} & \textbf{62.6} & \multicolumn{1}{c}{\textbf{64.0}} & \multicolumn{1}{c}{\textbf{94.7}} & \textbf{70.9} & \multicolumn{1}{c}{\textbf{50.4}} & \multicolumn{1}{c}{\textbf{93.7}} & \textbf{54.4} \\ \hline
\multicolumn{1}{|c|}{}                             & Zero-shot & \multicolumn{1}{c}{32.0} & \multicolumn{1}{c}{68.1} & 45.4 & \multicolumn{1}{c}{41.5} & \multicolumn{1}{c}{76.9}                         & 51.6 & \multicolumn{1}{c}{37.4} & \multicolumn{1}{c}{69.8} & 48.9 & \multicolumn{1}{c}{45.8} & \multicolumn{1}{c}{81.3} & 57.4 & \multicolumn{1}{c}{34.2} & \multicolumn{1}{c}{75.8} & 43.8 \\  
\multicolumn{1}{|c|}{}                             & 1-shot    & \multicolumn{1}{c}{47.7} & \multicolumn{1}{c}{86.6} & 54.9 & \multicolumn{1}{c}{47.8} & \multicolumn{1}{c}{82.0}                         & 57.0 & \multicolumn{1}{c}{43.4} & \multicolumn{1}{c}{76.6} & 0.54.1 & \multicolumn{1}{c}{51.9} & \multicolumn{1}{c}{85.6} & 62.2 & \multicolumn{1}{c}{42.0} & \multicolumn{1}{c}{85.5} & 477 \\  
\multicolumn{1}{|c|}{\multirow{-3}{*}{FMFruit-L}} & Fine-tuning & \multicolumn{1}{c}{57.5} & \multicolumn{1}{c}{93.6} & 62.7 & \multicolumn{1}{c}{59.8} & \multicolumn{1}{c}{91.5}                         & 0.662 & \multicolumn{1}{c}{56.5} & \multicolumn{1}{c}{88.5} & 0.634 & \multicolumn{1}{c}{63.6} & \multicolumn{1}{c}{93.6} & 71.1 & \multicolumn{1}{c}{49.3} & \multicolumn{1}{c}{93.7} & 53.2 \\ \hline
\end{tabular}
}
\end{table*}

\begin{figure*}[!ht]
  \centering  \includegraphics[width=0.99\textwidth]{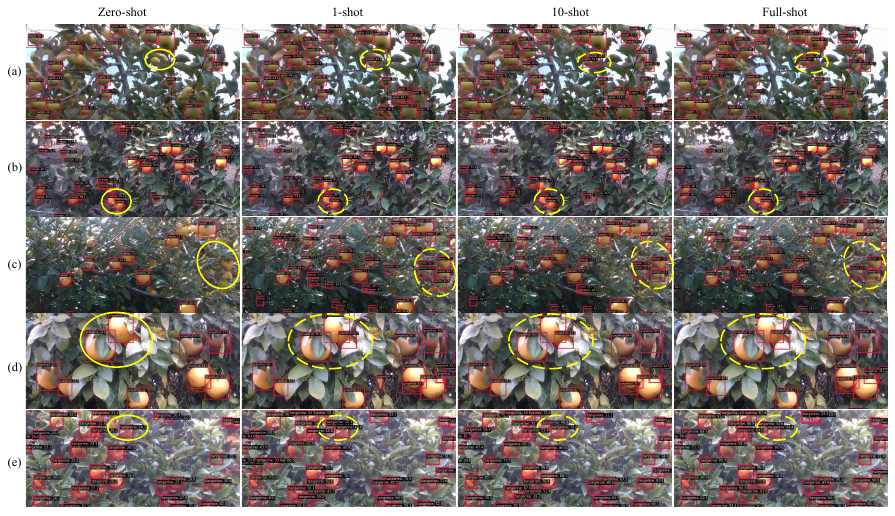}
  \caption{Zero-shot and few-shot fruit detection visualization examples for (a) apple, (b) orange, (c) lemon, (d) grapefruit, and (e) tangerine. The bounding box confidence threshold is set as 0.2 and 0.3 for zero-shot and few-shot, respectively. Best view via zoom in.}
  \label{fig:fruit_vis}
\end{figure*}

Figure~\ref{fig:fruit_vis} presents examples of detection outputs achieved by the FMFruit model under various few-shot configurations. These visualizations underscore FMFruit's robust open-set detection capabilities, particularly in zero-shot settings where the model undergoes no fine-tuning. This illustrates FMFruit's inherent ability to generalize and accurately identify fruits even without prior exposure to specific fruit class data. It is noted that in certain cases, such as with lemons, the model under the zero-shot setting misses some fruit instances due to occlusion and the small size of the fruits, as illustrated in Figure~\ref{fig:fruit_vis} (c). However, the model's detection capabilities are significantly enhanced through fine-tuning with just one single image (i.e., 1-shot), by effectively adapting to address challenges associated with fruit occlusion and small size. 

\subsection{Performance of cross-class generalization}
In this subsection, we evaluate the cross-class generalization capability of FMFruit to assess the impact of training on existing fruit classes on the detection performance of an unseen fruit class. Specifically, in this evaluation, the model is first trained on four fruit classes and subsequently tested on the fifth, unseen class. For instance, to test the model's generalization capability to detect lemons with cross-variety training data of other fruits, the model is first fine-tuned using data from oranges, apples, grapefruits, and tangerines, and then tested for its ability to detect lemons, a class not seen during training. This assessment helps us understand FMFruit's adaptability and effectiveness in recognizing new fruit types based on learned features from other fruit classes.

Table~\ref{tab:cross_generalization} summarizes the performance of FMFruit across three distinct training settings: zero-shot, where the model receives no training on any of the five fruit classes; cross-class, where the model is trained on four fruit classes and evaluated on the fifth, unseen class; and fine-tuning, where the model undergoes training on the specific fruit classes. The data results clearly demonstrate the efficacy of cross-class training in enhancing fruit detection capabilities. Specifically, cross-class training significantly boosts detection performance by 98.9, with an AP50 improvement from 45.7 to 90.9, nearly matching the performance in the fine-tuning setting, which achieves an AP50 of 92.7. This outcome underscores the potential of cross-class training to effectively prepare models for recognizing new fruit types.

\begin{table*}[!ht]
\renewcommand{\arraystretch}{1.4}
\centering
\caption{Cross-class generalization performance. In the cross-class training setting, the model is trained using four out of five fruit types, and its performance is then evaluated on a fifth, distinct fruit type that is not included in the training set.}
\label{tab:cross_generalization}
\resizebox{0.95 \textwidth}{!}{%
\begin{tabular}{|c|ccc|ccc|ccc|ccc|ccc|}
\hline
                     & \multicolumn{3}{c|}{Apple}                                      & \multicolumn{3}{c|}{Orange}                                                             & \multicolumn{3}{c|}{Lemon}                                      & \multicolumn{3}{c|}{Grapefruit}                                 & \multicolumn{3}{c|}{Tangerine}                                  \\  
\multirow{-2}{*}{}   & \multicolumn{1}{c}{mAP}   & \multicolumn{1}{c}{AP50}  & mAR   & \multicolumn{1}{c}{mAP}   & \multicolumn{1}{c}{AP50}  & mAR                           & \multicolumn{1}{c}{mAP}   & \multicolumn{1}{c}{AP50}  & mAR   & \multicolumn{1}{c}{mAP}   & \multicolumn{1}{c}{AP50}  & mAR   & \multicolumn{1}{c}{mAP}   & \multicolumn{1}{c}{AP50}  & mAR   \\ \hline \hline
Zero-shot            & \multicolumn{1}{c}{24.1} & \multicolumn{1}{c}{45.7} & 46.1 & \multicolumn{1}{c}{36.6} & \multicolumn{1}{c}{68.9} & \cellcolor[HTML]{FFFFFF}52.2 & \multicolumn{1}{c}{29.4} & \multicolumn{1}{c}{52.1} & 47.3 & \multicolumn{1}{c}{37.2} & \multicolumn{1}{c}{64.4} & 52.7 & \multicolumn{1}{c}{29.6} & \multicolumn{1}{c}{60.3} & 43.8 \\ 
Cross-class training & \multicolumn{1}{c}{53.0}  & \multicolumn{1}{c}{90.2} & 59.4 & \multicolumn{1}{c}{58.0}  & \multicolumn{1}{c}{89.4} & 64.3                         & \multicolumn{1}{c}{52.0}  & \multicolumn{1}{c}{83.8} & 59.6 & \multicolumn{1}{c}{60.8} & \multicolumn{1}{c}{90.1} & 71.1 & \multicolumn{1}{c}{47.8} & \multicolumn{1}{c}{92.2} & 52.0  \\ 
Fine-tuning   & \multicolumn{1}{c}{59.4} & \multicolumn{1}{c}{94.1} & 64.7 & \multicolumn{1}{c}{60.1} & \multicolumn{1}{c}{92.0} & 66.5                         & \multicolumn{1}{c}{56.0} & \multicolumn{1}{c}{88.0} & 62.6 & \multicolumn{1}{c}{64.0} & \multicolumn{1}{c}{94.7} & 70.9 & \multicolumn{1}{c}{50.4} & \multicolumn{1}{c}{93.7} & 54.4 \\ \hline
\end{tabular}
}
\end{table*}

\subsection{Performance on other fruit datasets}
In this subsection, we evaluate the performance of our model, FMFruit-T, across several established fruit datasets, specifically targeting the DeepBlueberry \citep{gonzalez2019deepblueberry}, StrawDI$\_$Db1 \citep{perez2020fast}, and MinneApple \citep{hani2020minneapple}. These datasets contain 125, 300, and 331 test images, respectively, with corresponding training sets comprising 184, 2800, and 670 images. This evaluation allows us to rigorously assess FMFruit-T's effectiveness across diverse fruit classes and scenarios.

\begin{figure*}[!ht]
    \centering
    \subfloat[Deepblueberry]{%
        \includegraphics[width=0.75\linewidth]{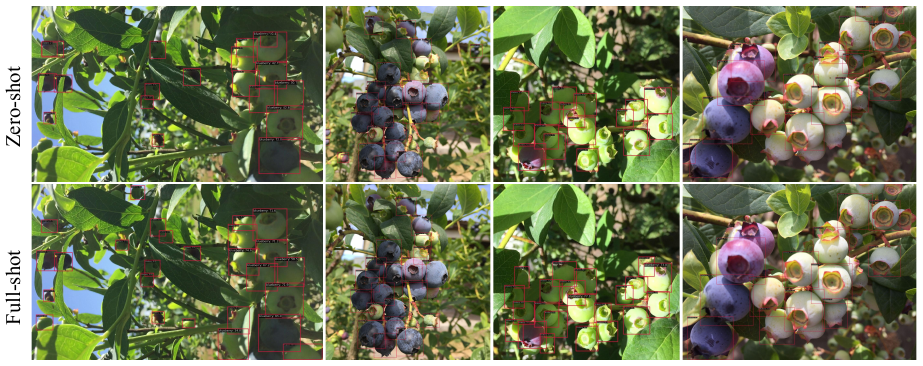}
        \label{fig:sub_blueberry}%
    }\\ 
    \vspace{-7pt}
    \subfloat[StrawberryDI]{%
        \includegraphics[width=0.75\linewidth]{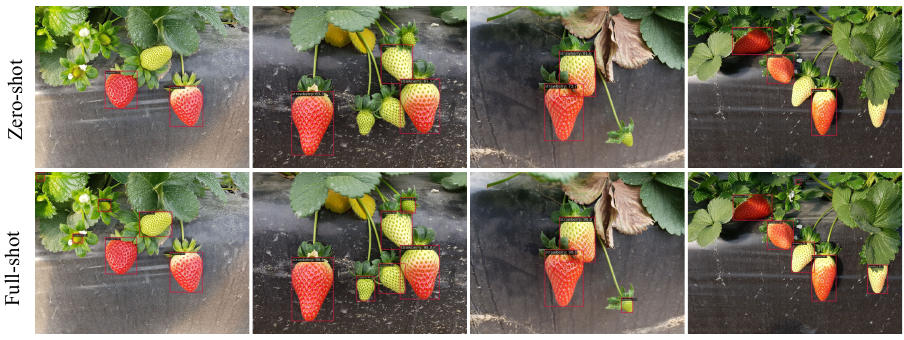}
        \label{fig:sub_strawberry}%
    }\\ 
    \vspace{-7pt}
    \subfloat[Minneappple]{%
        \includegraphics[width=0.75\linewidth]{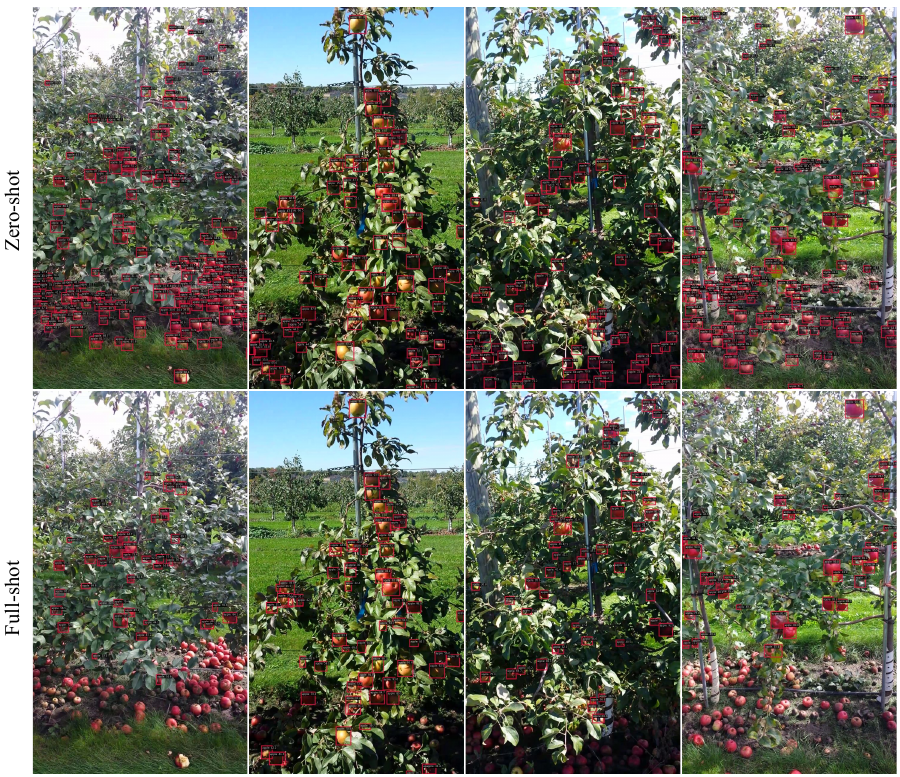}
        \label{fig: sub_apple}%
    }
    \caption{Fruit visualization results of zero-shot and fine-tuning on other public datasets, where the bounding box confidence threshold is set as 0.3. Best view via zoom in.}
    \label{fig:other_dataset}
\end{figure*}

Table \ref{tab:other_datasets} illustrates FMFruit-T's exemplary performance across all tested datasets. Remarkably, after pretraining on the MetaFruit dataset, FMFruit-T achieves mAP rates of 30.8, 47.6, and 20.7 on the DeepBlueberry, StrawDI$\_$Db1, and MinniApple datasets, respectively. Despite the absence of strawberry and blueberry classes in the MetaFruit dataset, FMFruit-T demonstrates a notable ability to generalize and adapt to these fruit types, even though it was not directly trained on them. The Minneapple dataset's inclusion of complex orchard environments similar to those in the MetaFruit dataset, along with numerous apples fallen on the ground (as shown in Figure~\ref{fig:other_dataset}), presents unique challenges, resulting in the lower initial detection accuracy in zero-shot settings. However, with an increased number of training samples, FMFruit-T adeptly distinguishes and focuses on apples located on trees, effectively ignoring those on the ground. Upon fine-tuning setting, FMFruit-T impressively attains AP scores of 69.4, 87.7, and 51.8 for the DeepBlueberry, StrawDI$\_$Db1, and MinniApple datasets, respectively, demonstrating its robustness and adaptability across diverse fruit detection scenarios.


\begin{table*}[!ht]
\renewcommand{\arraystretch}{1.4}
\centering
\caption{Performance on some of the existing fruit data: DeepBlueberry \citep{gonzalez2019deepblueberry}, StrawDI$\_$Db1 \citep{perez2020fast}, and MinneApple \citep{hani2020minneapple}.}
\label{tab:other_datasets}
\resizebox{0.65 \textwidth}{!}{%
\begin{tabular}{|c|ccc|ccc|ccc|}
\hline
                                & \multicolumn{3}{c|}{DeepBlueberry}                                  & \multicolumn{3}{c|}{StrawDI$\_$Db1}                                                         & \multicolumn{3}{c|}{Minneapple}                                      \\ 
\multirow{-2}{*}{}              & \multicolumn{1}{c}{mAP}   & \multicolumn{1}{c}{AP50}  & mAR   & \multicolumn{1}{c}{mAP}   & \multicolumn{1}{c}{AP50}  & mAR                           & \multicolumn{1}{c}{mAP}   & \multicolumn{1}{c}{AP50}  & mAR   \\ \hline \hline
Zero-shot(Pre-t rained on MetaFruit) & \multicolumn{1}{c}{30.8} & \multicolumn{1}{c}{50.4} & 51.0 & \multicolumn{1}{c}{47.6} & \multicolumn{1}{c}{60.3} & \cellcolor[HTML]{FFFFFF}69.7 & \multicolumn{1}{c}{20.7} & \multicolumn{1}{c}{40.4} & 37.5 \\ 
1-shot                          & \multicolumn{1}{c}{47.8} & \multicolumn{1}{c}{65.7} & 65.6 & \multicolumn{1}{c}{64.3} & \multicolumn{1}{c}{76.8} & 80.9                         & \multicolumn{1}{c}{18.6} & \multicolumn{1}{c}{34.4} & 46.9 \\ 
5-shot                          & \multicolumn{1}{c}{47.8} & \multicolumn{1}{c}{67.0} & 67.3 & \multicolumn{1}{c}{79.5} & \multicolumn{1}{c}{91.3} & 84.1                         & \multicolumn{1}{c}{17.9} & \multicolumn{1}{c}{32.3} & 46.7 \\ 
10-shot                         & \multicolumn{1}{c}{55.4} & \multicolumn{1}{c}{74.5} & 69.2 & \multicolumn{1}{c}{80.7} & \multicolumn{1}{c}{92.6} & 85.9                         & \multicolumn{1}{c}{36.4} & \multicolumn{1}{c}{62.2} & 53.0 \\ 
Fine-tuning                       & \multicolumn{1}{c}{69.4} & \multicolumn{1}{c}{90.1} & 76.9 & \multicolumn{1}{c}{87.7} & \multicolumn{1}{c}{97.7} & 89.8                         & \multicolumn{1}{c}{51.8} & \multicolumn{1}{c}{86.1} & 61.3 \\ \hline
\end{tabular}
}
\end{table*}

\subsection{Performance of referring expression comprehension (REC)}\label{sec:hri}
In this subsection, we present an initial evaluation of our FMFruit model's ability in terms of REC. The model is tasked with processing human instructions provided in natural language, identifying the critical elements of these instructions, and selecting features that accurately correspond to the described text. 

\begin{figure*}[!ht]
  \centering
  \includegraphics[width=0.8\textwidth]{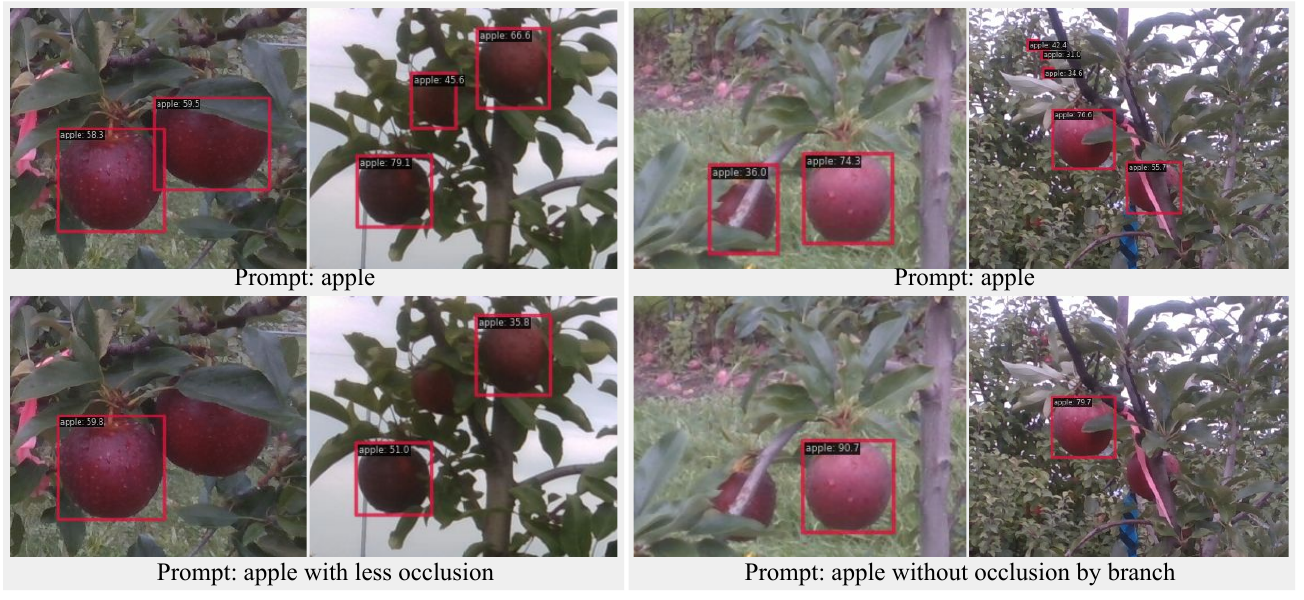}
  \caption{Visualization examples of referring object detection. The first row displays results for the prompt ``apple'', while the second row shows responses to a more specific prompt, such as ``apple with less occlusion'' or ``apple without occlusion by branch''.}
  \label{fig:rec}
\end{figure*}

Figure~\ref{fig:rec} shows the REC results. The first illustrative set involves the model detecting apples with minimal occlusion, guided by the specific instruction ``apple with less occlusion''. FMFruit demonstrates proficiency in accurately isolating and excluding apples that are heavily occluded by leaves, adhering closely to the given instructions. The second example demonstrates the model's ability to filter out apples occluded by branches, following the instruction ``apple without occlusion by branch''. Unsurprisingly, FMFruit exhibits exceptional adaptability by focusing detection on apples without branch occlusion. These scenarios highlight FMFruit's precise interpretation and execution based on specific linguistic instructions, underscoring its sophisticated ability to utilize referring expressions for enhanced fruit detection accuracy.

\begin{table}[!ht]
\renewcommand{\arraystretch}{1.4}
\centering
\caption{Model inference time. It is accessed on a single RTX 2080Ti GPU.}
\label{tab:inference_time}
\resizebox{0.43 \textwidth}{!}{%
\begin{tabular}{|c|c|c|}
\hline
Model       & \multicolumn{1}{c|}{FPS (imgs/s)} & Inference time per image (ms) \\ \hline \hline
Retinanet   & 21.9                              & 45.7                           \\ 
Faster RCNN & 20.2                              & 49.5                           \\ 
FCOS        & 18.9                              & 52.9                           \\ 
RTMDet      & 53.2                              & 18.8                           \\ \hline
FMFruit-T   & 5.5                               & 181.8                          \\ 
FMFruit-L   & 3.9                               & 256.4                          \\ \hline
\end{tabular}
}
\end{table}

\section{Discussion}
\label{sec:diss}
Fruit detection is a widely studied research topic but is still a practical challenge. Traditional DL methods have shown considerable success, yet they tend to be specialized for certain fruit types and specific scenarios, limiting their applicability to new orchard environments and different fruit classes. In response to this limitation, our study delves into the potential of VFMs to tackle a wider range of fruit detection challenges. Additionally, we introduce the MetaFruit dataset, encompassing 248,015 labeled instances across five fruit classes, to support and enhance the development and evaluation of advanced fruit detection models. Despite its contributions, this study acknowledges certain limitations, as elaborated below.

\subsection{Challenges in real-world implementation}
Implementing FMs in agricultural applications comes with some challenges, particularly regarding inference speed and model size which often require significant computing resources \citep{bommasani2021opportunities}. As shown in Table~\ref{tab:inference_time}, our proposed FMFruit models have the largest inference time, which could limit the deployment of FMs in many on-field agricultural settings, as the downstream tasks often require immediate action based on the model's outputs. For example, after outputting the fruit location, the fruit-picking system needs to implement other actions immediately, such as decision-making and path planning. In addition, the complexity and size of FMs demand large computing resources and memory bandwidth, which may not be practical for real-world deployment.

To overcome these challenges, recent research has focused on the model optimization techniques \citep{zhu2023survey} and made great progress. For instance, model compression \citep{cheng2017survey, choudhary2020comprehensive} can significantly reduce the model size and speed up the inference without compromising performance. These techniques include quantization, knowledge distillation, and pruning, among others, each contributing to more efficient deployment of FMs in resource-constrained environments like agriculture.  For example, SqueezeLLM \citep{kim2023squeezellm} proposed a post-training
quantization framework to enable lossless compression and achieve higher quantization performance under the same memory constraint. 

Furthermore, adopting edge computing strategies can accelerate the inference process by facilitating data processing near its point of origin \citep{chen2019deep}. In the context of agriculture, edge devices such as drones or field sensors can process data on-site, allowing for immediate decision-making without dependence on remote servers.  An illustrative example of this approach is MobileSAM \citep{zhang2023faster}, which is designed by distilling the knowledge from the heavy image encoder (ViT-H in the original SAM \citep{kirillov2023segment}) to a lightweight image encoder, enable the model implementation in mobile devices. Specifically, the MobileSAM achieves an inference time of 12ms and a model parameter of 9.66M, compared with the original SAM  with an inference time of 456ms and a model parameter of 615M.

\subsection{Integration of LLMs}
The realm of Large Language Models (LLMs) and Foundation Models (FMs) has seen remarkable advancements, finding applications in diverse fields including ChatGPT \citep{achiam2023gpt}, robotics \citep{firoozi2023foundation}, and agriculture \citep{yang2023sam}. The preliminary investigations into the use of LLMs and FMs within agricultural contexts reveal significant promise for their integration into farming technologies, suggesting a fruitful avenue for enhancing agricultural practices through advanced computational models. In Section~\ref{sec:hri}, we explore the efficacy of Referring Expression Comprehension by leveraging human instructions to refine detection outcomes. To seamlessly integrate language and visual modalities, we employ a language-guided query selection method \citep{liu2023grounding}, which selects features that closely align with the input text, utilizing the principles of the grounded language-image pre-training (GLIP) model \citep{li2022grounded}. This method promises more precise and contextually relevant detection capabilities. However, it necessitates the preparation of well-organized and labeled (image, text) pairs for training, a process that is both time-intensive and complex. Looking ahead, the exploration of integrating mature LLM and FM developer Application Programming Interfaces (APIs) presents an exciting avenue. For instance, OpenAI has made their ChatGPT API\footnote{ChatGPT API:\url{https://openai.com/blog/introducing-chatgpt-and-whisper-apis}.} available to the public, enabling researchers to develop their own applications and tools leveraging these advanced platforms. 

Human-robot interaction (HRI), a multidisciplinary field that studies how humans and robots interact, presents another opportunity for integrating LLMs and FMs into fruit-harvesting robots \citep{wang2024large}.  With the power of LLMs and FMs, fruit-harvesting robots can be endowed with enhanced comprehension abilities, enabling them to understand and execute complex instructions provided by humans in natural language. This integration not only facilitates smoother and more intuitive communication between humans and robots but also significantly improves the robots' adaptability and decision-making capabilities in dynamic orchard environments.

\section{Summary}
\label{sec:conclusion}
Fruit detection is a pivotal component in the development of robotic fruit harvesting systems. Central to successful fruit detection is the assembly of a substantial, accurately labeled fruit dataset and the subsequent development of robust DL models. This paper introduces, to date, the most extensive fruit detection dataset pertinent to U.S. commercial orchards, encompassing 4,248 images across 5 fruit classes, annotated with a total of 248,015 bounding boxes, gathered under diverse natural field lighting conditions and different geographic locations. Moreover, we have developed an innovative open-set fruit detection system that utilizes the advanced capabilities of VFMs to identify a wide range of fruits. This model shows superior capability of detecting unknown fruits and can achieve fine performance under zero-shot and few-shot learning scenarios. Furthermore, the model demonstrates cross-class generalization capabilities by being trained on known fruit classes and then tested on novel classes, showcasing its exceptional open-set detection ability. Lastly, the model shows superior ability in language referring expression comprehension, thus providing opportunities for human-robot interactions. The fruit detection dataset and source codes for model development and evaluation are now publicly accessible to the research community.




\typeout{}
\bibliography{ref}

\begin{thebibliography}{84}
\providecommand{\natexlab}[1]{#1}
\providecommand{\url}[1]{\texttt{#1}}
\expandafter\ifx\csname urlstyle\endcsname\relax
  \providecommand{\doi}[1]{doi: #1}\else
  \providecommand{\doi}{doi: \begingroup \urlstyle{rm}\Url}\fi

\bibitem[Achiam et~al.(2023)Achiam, Adler, Agarwal, Ahmad, Akkaya, Aleman, Almeida, Altenschmidt, Altman, Anadkat, et~al.]{achiam2023gpt}
J.~Achiam, S.~Adler, S.~Agarwal, L.~Ahmad, I.~Akkaya, F.~L. Aleman, D.~Almeida, J.~Altenschmidt, S.~Altman, S.~Anadkat, et~al.
\newblock Gpt-4 technical report.
\newblock \emph{arXiv preprint arXiv:2303.08774}, 2023.

\bibitem[Bargoti and Underwood(2017)]{bargoti2017deep}
S.~Bargoti and J.~Underwood.
\newblock Deep fruit detection in orchards.
\newblock In \emph{2017 IEEE international conference on robotics and automation (ICRA)}, pages 3626--3633. IEEE, 2017.

\bibitem[Bhusal et~al.(2019)Bhusal, Karkee, and Zhang]{bhusal2019apple}
S.~Bhusal, M.~Karkee, and Q.~Zhang.
\newblock Apple dataset benchmark from orchard environment in modern fruiting wall.
\newblock 2019.

\bibitem[Bommasani et~al.(2021)Bommasani, Hudson, Adeli, Altman, Arora, von Arx, Bernstein, Bohg, Bosselut, Brunskill, et~al.]{bommasani2021opportunities}
R.~Bommasani, D.~A. Hudson, E.~Adeli, R.~Altman, S.~Arora, S.~von Arx, M.~S. Bernstein, J.~Bohg, A.~Bosselut, E.~Brunskill, et~al.
\newblock On the opportunities and risks of foundation models.
\newblock \emph{arXiv preprint arXiv:2108.07258}, 2021.

\bibitem[Chaivivatrakul and Dailey(2014)]{chaivivatrakul2014texture}
S.~Chaivivatrakul and M.~N. Dailey.
\newblock Texture-based fruit detection.
\newblock \emph{Precision Agriculture}, 15:\penalty0 662--683, 2014.

\bibitem[Chen et~al.(2024)Chen, Qi, Zheng, Lu, Huang, and Li]{chen2024synthetic}
D.~Chen, X.~Qi, Y.~Zheng, Y.~Lu, Y.~Huang, and Z.~Li.
\newblock Synthetic data augmentation by diffusion probabilistic models to enhance weed recognition.
\newblock \emph{Computers and Electronics in Agriculture}, 216:\penalty0 108517, 2024.

\bibitem[Chen and Ran(2019)]{chen2019deep}
J.~Chen and X.~Ran.
\newblock Deep learning with edge computing: A review.
\newblock \emph{Proceedings of the IEEE}, 107\penalty0 (8):\penalty0 1655--1674, 2019.

\bibitem[Cheng et~al.(2017)Cheng, Wang, Zhou, and Zhang]{cheng2017survey}
Y.~Cheng, D.~Wang, P.~Zhou, and T.~Zhang.
\newblock A survey of model compression and acceleration for deep neural networks.
\newblock \emph{arXiv preprint arXiv:1710.09282}, 2017.

\bibitem[Choudhary et~al.(2020)Choudhary, Mishra, Goswami, and Sarangapani]{choudhary2020comprehensive}
T.~Choudhary, V.~Mishra, A.~Goswami, and J.~Sarangapani.
\newblock A comprehensive survey on model compression and acceleration.
\newblock \emph{Artificial Intelligence Review}, 53:\penalty0 5113--5155, 2020.

\bibitem[Chu et~al.(2021)Chu, Li, Lammers, Lu, and Liu]{chu2021deep}
P.~Chu, Z.~Li, K.~Lammers, R.~Lu, and X.~Liu.
\newblock Deep learning-based apple detection using a suppression mask r-cnn.
\newblock \emph{Pattern Recognition Letters}, 147:\penalty0 206--211, 2021.

\bibitem[Chu et~al.(2023)Chu, Li, Zhang, Chen, Lammers, and Lu]{chu2023o2rnet}
P.~Chu, Z.~Li, K.~Zhang, D.~Chen, K.~Lammers, and R.~Lu.
\newblock O2rnet: Occluder-occludee relational network for robust apple detection in clustered orchard environments.
\newblock \emph{Smart Agricultural Technology}, page 100284, 2023.
\newblock ISSN 2772-3755.
\newblock \doi{https://doi.org/10.1016/j.atech.2023.100284}.
\newblock URL \url{https://www.sciencedirect.com/science/article/pii/S2772375523001132}.

\bibitem[Dang et~al.(2023)Dang, Chen, Lu, and Li]{dang2023yoloweeds}
F.~Dang, D.~Chen, Y.~Lu, and Z.~Li.
\newblock Yoloweeds: a novel benchmark of yolo object detectors for multi-class weed detection in cotton production systems.
\newblock \emph{Computers and Electronics in Agriculture}, 205:\penalty0 107655, 2023.

\bibitem[Devlin et~al.(2018)Devlin, Chang, Lee, and Toutanova]{devlin2018bert}
J.~Devlin, M.-W. Chang, K.~Lee, and K.~Toutanova.
\newblock Bert: Pre-training of deep bidirectional transformers for language understanding.
\newblock \emph{arXiv preprint arXiv:1810.04805}, 2018.

\bibitem[Firoozi et~al.(2023)Firoozi, Tucker, Tian, Majumdar, Sun, Liu, Zhu, Song, Kapoor, Hausman, et~al.]{firoozi2023foundation}
R.~Firoozi, J.~Tucker, S.~Tian, A.~Majumdar, J.~Sun, W.~Liu, Y.~Zhu, S.~Song, A.~Kapoor, K.~Hausman, et~al.
\newblock Foundation models in robotics: Applications, challenges, and the future.
\newblock \emph{arXiv preprint arXiv:2312.07843}, 2023.

\bibitem[Fu et~al.(2018)Fu, Feng, Majeed, Zhang, Zhang, Karkee, and Zhang]{fu2018kiwifruit}
L.~Fu, Y.~Feng, Y.~Majeed, X.~Zhang, J.~Zhang, M.~Karkee, and Q.~Zhang.
\newblock Kiwifruit detection in field images using faster r-cnn with zfnet.
\newblock \emph{IFAC-PapersOnLine}, 51\penalty0 (17):\penalty0 45--50, 2018.

\bibitem[Fu et~al.(2020)Fu, Majeed, Zhang, Karkee, and Zhang]{fu2020faster}
L.~Fu, Y.~Majeed, X.~Zhang, M.~Karkee, and Q.~Zhang.
\newblock Faster r--cnn--based apple detection in dense-foliage fruiting-wall trees using rgb and depth features for robotic harvesting.
\newblock \emph{Biosystems Engineering}, 197:\penalty0 245--256, 2020.

\bibitem[Gai et~al.(2023)Gai, Chen, and Yuan]{gai2023detection}
R.~Gai, N.~Chen, and H.~Yuan.
\newblock A detection algorithm for cherry fruits based on the improved yolo-v4 model.
\newblock \emph{Neural Computing and Applications}, 35\penalty0 (19):\penalty0 13895--13906, 2023.

\bibitem[Gao et~al.(2020)Gao, Fu, Zhang, Majeed, Li, Karkee, and Zhang]{gao2020multi}
F.~Gao, L.~Fu, X.~Zhang, Y.~Majeed, R.~Li, M.~Karkee, and Q.~Zhang.
\newblock Multi-class fruit-on-plant detection for apple in snap system using faster r-cnn.
\newblock \emph{Computers and Electronics in Agriculture}, 176:\penalty0 105634, 2020.

\bibitem[Gen{\'e}-Mola et~al.(2019)Gen{\'e}-Mola, Vilaplana, Rosell-Polo, Morros, Ruiz-Hidalgo, and Gregorio]{gene2019kfuji}
J.~Gen{\'e}-Mola, V.~Vilaplana, J.~R. Rosell-Polo, J.-R. Morros, J.~Ruiz-Hidalgo, and E.~Gregorio.
\newblock Kfuji rgb-ds database: Fuji apple multi-modal images for fruit detection with color, depth and range-corrected ir data.
\newblock \emph{Data in brief}, 25:\penalty0 104289, 2019.

\bibitem[Gen{\'e}-Mola et~al.(2020{\natexlab{a}})Gen{\'e}-Mola, Gregorio, Cheein, Guevara, Llorens, Sanz-Cortiella, Escola, and Rosell-Polo]{gene2020lfuji}
J.~Gen{\'e}-Mola, E.~Gregorio, F.~A. Cheein, J.~Guevara, J.~Llorens, R.~Sanz-Cortiella, A.~Escola, and J.~R. Rosell-Polo.
\newblock Lfuji-air dataset: Annotated 3d lidar point clouds of fuji apple trees for fruit detection scanned under different forced air flow conditions.
\newblock \emph{Data in brief}, 29:\penalty0 105248, 2020{\natexlab{a}}.

\bibitem[Gen{\'e}-Mola et~al.(2020{\natexlab{b}})Gen{\'e}-Mola, Sanz-Cortiella, Rosell-Polo, Morros, Ruiz-Hidalgo, Vilaplana, and Gregorio]{gene2020fuji}
J.~Gen{\'e}-Mola, R.~Sanz-Cortiella, J.~R. Rosell-Polo, J.-R. Morros, J.~Ruiz-Hidalgo, V.~Vilaplana, and E.~Gregorio.
\newblock Fuji-sfm dataset: A collection of annotated images and point clouds for fuji apple detection and location using structure-from-motion photogrammetry.
\newblock \emph{Data in brief}, 30:\penalty0 105591, 2020{\natexlab{b}}.

\bibitem[Geng et~al.(2020)Geng, Huang, and Chen]{geng2020recent}
C.~Geng, S.-j. Huang, and S.~Chen.
\newblock Recent advances in open set recognition: A survey.
\newblock \emph{IEEE transactions on pattern analysis and machine intelligence}, 43\penalty0 (10):\penalty0 3614--3631, 2020.

\bibitem[Gené-Mola et~al.(2020)Gené-Mola, Gregorio, {Auat Cheein}, Guevara, Llorens, Sanz-Cortiella, Escolà, and Rosell-Polo]{GENEMOLA2020105248}
J.~Gené-Mola, E.~Gregorio, F.~{Auat Cheein}, J.~Guevara, J.~Llorens, R.~Sanz-Cortiella, A.~Escolà, and J.~R. Rosell-Polo.
\newblock Lfuji-air dataset: Annotated 3d lidar point clouds of fuji apple trees for fruit detection scanned under different forced air flow conditions.
\newblock \emph{Data in Brief}, 29:\penalty0 105248, 2020.
\newblock ISSN 2352-3409.
\newblock \doi{https://doi.org/10.1016/j.dib.2020.105248}.
\newblock URL \url{https://www.sciencedirect.com/science/article/pii/S2352340920301426}.

\bibitem[Girshick(2015)]{girshick2015fast}
R.~Girshick.
\newblock Fast r-cnn.
\newblock In \emph{Proceedings of the IEEE international conference on computer vision}, pages 1440--1448, 2015.

\bibitem[Gongal et~al.(2015)Gongal, Amatya, Karkee, Zhang, and Lewis]{gongal2015sensors}
A.~Gongal, S.~Amatya, M.~Karkee, Q.~Zhang, and K.~Lewis.
\newblock Sensors and systems for fruit detection and localization: A review.
\newblock \emph{Computers and Electronics in Agriculture}, 116:\penalty0 8--19, 2015.

\bibitem[Gonzalez et~al.(2019)Gonzalez, Arellano, and Tapia]{gonzalez2019deepblueberry}
S.~Gonzalez, C.~Arellano, and J.~E. Tapia.
\newblock Deepblueberry: Quantification of blueberries in the wild using instance segmentation.
\newblock \emph{Ieee Access}, 7:\penalty0 105776--105788, 2019.

\bibitem[H{\"a}ni et~al.(2020)H{\"a}ni, Roy, and Isler]{hani2020minneapple}
N.~H{\"a}ni, P.~Roy, and V.~Isler.
\newblock Minneapple: a benchmark dataset for apple detection and segmentation.
\newblock \emph{IEEE Robotics and Automation Letters}, 5\penalty0 (2):\penalty0 852--858, 2020.

\bibitem[He et~al.(2017)He, Gkioxari, Doll{\'a}r, and Girshick]{he2017mask}
K.~He, G.~Gkioxari, P.~Doll{\'a}r, and R.~Girshick.
\newblock Mask r-cnn.
\newblock In \emph{Proceedings of the IEEE international conference on computer vision}, pages 2961--2969, 2017.

\bibitem[Kamilaris and Prenafeta-Bold{\'u}(2018)]{kamilaris2018deep}
A.~Kamilaris and F.~X. Prenafeta-Bold{\'u}.
\newblock Deep learning in agriculture: A survey.
\newblock \emph{Computers and electronics in agriculture}, 147:\penalty0 70--90, 2018.

\bibitem[Kazemzadeh et~al.(2014)Kazemzadeh, Ordonez, Matten, and Berg]{kazemzadeh2014referitgame}
S.~Kazemzadeh, V.~Ordonez, M.~Matten, and T.~Berg.
\newblock Referitgame: Referring to objects in photographs of natural scenes.
\newblock In \emph{Proceedings of the 2014 conference on empirical methods in natural language processing (EMNLP)}, pages 787--798, 2014.

\bibitem[Kestur et~al.(2019)Kestur, Meduri, and Narasipura]{kestur2019mangonet}
R.~Kestur, A.~Meduri, and O.~Narasipura.
\newblock Mangonet: A deep semantic segmentation architecture for a method to detect and count mangoes in an open orchard.
\newblock \emph{Engineering Applications of Artificial Intelligence}, 77:\penalty0 59--69, 2019.

\bibitem[Kim et~al.(2023)Kim, Hooper, Gholami, Dong, Li, Shen, Mahoney, and Keutzer]{kim2023squeezellm}
S.~Kim, C.~Hooper, A.~Gholami, Z.~Dong, X.~Li, S.~Shen, M.~Mahoney, and K.~Keutzer.
\newblock Squeezellm: Dense-and-sparse quantization.
\newblock \emph{arXiv}, 2023.

\bibitem[Kirillov et~al.(2023)Kirillov, Mintun, Ravi, Mao, Rolland, Gustafson, Xiao, Whitehead, Berg, Lo, et~al.]{kirillov2023segment}
A.~Kirillov, E.~Mintun, N.~Ravi, H.~Mao, C.~Rolland, L.~Gustafson, T.~Xiao, S.~Whitehead, A.~C. Berg, W.-Y. Lo, et~al.
\newblock Segment anything.
\newblock In \emph{Proceedings of the IEEE/CVF International Conference on Computer Vision}, pages 4015--4026, 2023.

\bibitem[Koirala et~al.(2019)Koirala, Walsh, Wang, and McCarthy]{koirala2019deep}
A.~Koirala, K.~B. Walsh, Z.~Wang, and C.~McCarthy.
\newblock Deep learning for real-time fruit detection and orchard fruit load estimation: Benchmarking of ‘mangoyolo’.
\newblock \emph{Precision Agriculture}, 20\penalty0 (6):\penalty0 1107--1135, 2019.

\bibitem[Krasin et~al.(2017)Krasin, Duerig, Alldrin, Ferrari, Abu-El-Haija, Kuznetsova, Rom, Uijlings, Popov, Veit, et~al.]{krasin2017openimages}
I.~Krasin, T.~Duerig, N.~Alldrin, V.~Ferrari, S.~Abu-El-Haija, A.~Kuznetsova, H.~Rom, J.~Uijlings, S.~Popov, A.~Veit, et~al.
\newblock Openimages: A public dataset for large-scale multi-label and multi-class image classification.
\newblock \emph{Dataset available from https://github. com/openimages}, 2\penalty0 (3):\penalty0 18, 2017.

\bibitem[Krishna et~al.(2017)Krishna, Zhu, Groth, Johnson, Hata, Kravitz, Chen, Kalantidis, Li, Shamma, et~al.]{krishna2017visual}
R.~Krishna, Y.~Zhu, O.~Groth, J.~Johnson, K.~Hata, J.~Kravitz, S.~Chen, Y.~Kalantidis, L.-J. Li, D.~A. Shamma, et~al.
\newblock Visual genome: Connecting language and vision using crowdsourced dense image annotations.
\newblock \emph{International journal of computer vision}, 123:\penalty0 32--73, 2017.

\bibitem[LeCun et~al.(2015)LeCun, Bengio, and Hinton]{lecun2015deep}
Y.~LeCun, Y.~Bengio, and G.~Hinton.
\newblock Deep learning.
\newblock \emph{nature}, 521\penalty0 (7553):\penalty0 436--444, 2015.

\bibitem[Li et~al.(2023{\natexlab{a}})Li, Chen, Li, Huang, and Morris]{li2023ml}
J.~Li, D.~Chen, Z.~Li, Y.~Huang, and D.~Morris.
\newblock Ml/dl in agriculture through label-efficient learning.
\newblock In \emph{2023 ASABE Annual International Meeting}, page~1. American Society of Agricultural and Biological Engineers, 2023{\natexlab{a}}.

\bibitem[Li et~al.(2023{\natexlab{b}})Li, Chen, Qi, Li, Huang, Morris, and Tan]{li2023label}
J.~Li, D.~Chen, X.~Qi, Z.~Li, Y.~Huang, D.~Morris, and X.~Tan.
\newblock Label-efficient learning in agriculture: A comprehensive review.
\newblock \emph{Computers and Electronics in Agriculture}, 215:\penalty0 108412, 2023{\natexlab{b}}.

\bibitem[Li et~al.(2023{\natexlab{c}})Li, Xu, Xiang, Chen, Zhuang, Yin, and Li]{li2023foundation}
J.~Li, M.~Xu, L.~Xiang, D.~Chen, W.~Zhuang, X.~Yin, and Z.~Li.
\newblock Foundation models in smart agriculture: Basics, opportunities, and challenges.
\newblock \emph{arXiv preprint arXiv:2308.06668}, 2023{\natexlab{c}}.

\bibitem[Li et~al.(2024{\natexlab{a}})Li, Chen, Yin, and Li]{li2024performance}
J.~Li, D.~Chen, X.~Yin, and Z.~Li.
\newblock Performance evaluation of semi-supervised learning frameworks for multi-class weed detection.
\newblock \emph{arXiv preprint arXiv:2403.03390}, 2024{\natexlab{a}}.

\bibitem[Li et~al.(2024{\natexlab{b}})Li, Magar, Chen, Lin, Wang, Yin, Zhuang, and Li]{li2024soybeannet}
J.~Li, R.~T. Magar, D.~Chen, F.~Lin, D.~Wang, X.~Yin, W.~Zhuang, and Z.~Li.
\newblock Soybeannet: Transformer-based convolutional neural network for soybean pod counting from unmanned aerial vehicle (uav) images.
\newblock \emph{Computers and Electronics in Agriculture}, 220:\penalty0 108861, 2024{\natexlab{b}}.

\bibitem[Li et~al.(2022)Li, Zhang, Zhang, Yang, Li, Zhong, Wang, Yuan, Zhang, Hwang, et~al.]{li2022grounded}
L.~H. Li, P.~Zhang, H.~Zhang, J.~Yang, C.~Li, Y.~Zhong, L.~Wang, L.~Yuan, L.~Zhang, J.-N. Hwang, et~al.
\newblock Grounded language-image pre-training.
\newblock In \emph{Proceedings of the IEEE/CVF Conference on Computer Vision and Pattern Recognition}, pages 10965--10975, 2022.

\bibitem[Li et~al.(2011)Li, Lee, and Hsu]{li2011review}
P.~Li, S.-h. Lee, and H.-Y. Hsu.
\newblock Review on fruit harvesting method for potential use of automatic fruit harvesting systems.
\newblock \emph{Procedia Engineering}, 23:\penalty0 351--366, 2011.

\bibitem[Li et~al.(2024{\natexlab{c}})Li, Ma, Li, Zhang, Zhang, and Zhou]{li2024cotton}
Q.~Li, W.~Ma, H.~Li, X.~Zhang, R.~Zhang, and W.~Zhou.
\newblock Cotton-yolo: Improved yolov7 for rapid detection of foreign fibers in seed cotton.
\newblock \emph{Computers and Electronics in Agriculture}, 219:\penalty0 108752, 2024{\natexlab{c}}.

\bibitem[Lin et~al.(2014)Lin, Maire, Belongie, Hays, Perona, Ramanan, Doll{\'a}r, and Zitnick]{lin2014microsoft}
T.-Y. Lin, M.~Maire, S.~Belongie, J.~Hays, P.~Perona, D.~Ramanan, P.~Doll{\'a}r, and C.~L. Zitnick.
\newblock Microsoft coco: Common objects in context.
\newblock In \emph{Computer Vision--ECCV 2014: 13th European Conference, Zurich, Switzerland, September 6-12, 2014, Proceedings, Part V 13}, pages 740--755. Springer, 2014.

\bibitem[Lin et~al.(2017)Lin, Goyal, Girshick, He, and Doll{\'a}r]{lin2017focal}
T.-Y. Lin, P.~Goyal, R.~Girshick, K.~He, and P.~Doll{\'a}r.
\newblock Focal loss for dense object detection.
\newblock In \emph{Proceedings of the IEEE international conference on computer vision}, pages 2980--2988, 2017.

\bibitem[Liu et~al.(2023)Liu, Zeng, Ren, Li, Zhang, Yang, Li, Yang, Su, Zhu, et~al.]{liu2023grounding}
S.~Liu, Z.~Zeng, T.~Ren, F.~Li, H.~Zhang, J.~Yang, C.~Li, J.~Yang, H.~Su, J.~Zhu, et~al.
\newblock Grounding dino: Marrying dino with grounded pre-training for open-set object detection.
\newblock \emph{arXiv preprint arXiv:2303.05499}, 2023.

\bibitem[Liu et~al.(2021)Liu, Lin, Cao, Hu, Wei, Zhang, Lin, and Guo]{liu2021swin}
Z.~Liu, Y.~Lin, Y.~Cao, H.~Hu, Y.~Wei, Z.~Zhang, S.~Lin, and B.~Guo.
\newblock Swin transformer: Hierarchical vision transformer using shifted windows.
\newblock In \emph{Proceedings of the IEEE/CVF international conference on computer vision}, pages 10012--10022, 2021.

\bibitem[Lu and Young(2020)]{lu2020survey}
Y.~Lu and S.~Young.
\newblock A survey of public datasets for computer vision tasks in precision agriculture.
\newblock \emph{Computers and Electronics in Agriculture}, 178:\penalty0 105760, 2020.

\bibitem[Lyu et~al.(2022)Lyu, Zhang, Huang, Zhou, Wang, Liu, Zhang, and Chen]{lyu2022rtmdet}
C.~Lyu, W.~Zhang, H.~Huang, Y.~Zhou, Y.~Wang, Y.~Liu, S.~Zhang, and K.~Chen.
\newblock Rtmdet: An empirical study of designing real-time object detectors.
\newblock \emph{arXiv preprint arXiv:2212.07784}, 2022.

\bibitem[Meshram and Patil(2022)]{meshram2022fruitnet}
V.~Meshram and K.~Patil.
\newblock Fruitnet: Indian fruits image dataset with quality for machine learning applications.
\newblock \emph{Data in Brief}, 40:\penalty0 107686, 2022.

\bibitem[Mirhaji et~al.(2021)Mirhaji, Soleymani, Asakereh, and Mehdizadeh]{mirhaji2021fruit}
H.~Mirhaji, M.~Soleymani, A.~Asakereh, and S.~A. Mehdizadeh.
\newblock Fruit detection and load estimation of an orange orchard using the yolo models through simple approaches in different imaging and illumination conditions.
\newblock \emph{Computers and Electronics in Agriculture}, 191:\penalty0 106533, 2021.

\bibitem[Paszke et~al.(2019)Paszke, Gross, Massa, Lerer, Bradbury, Chanan, Killeen, Lin, Gimelshein, Antiga, et~al.]{paszke2019pytorch}
A.~Paszke, S.~Gross, F.~Massa, A.~Lerer, J.~Bradbury, G.~Chanan, T.~Killeen, Z.~Lin, N.~Gimelshein, L.~Antiga, et~al.
\newblock Pytorch: An imperative style, high-performance deep learning library.
\newblock \emph{Advances in neural information processing systems}, 32, 2019.

\bibitem[P{\'e}rez-Borrero et~al.(2020)P{\'e}rez-Borrero, Mar{\'\i}n-Santos, Gegundez-Arias, and Cort{\'e}s-Ancos]{perez2020fast}
I.~P{\'e}rez-Borrero, D.~Mar{\'\i}n-Santos, M.~E. Gegundez-Arias, and E.~Cort{\'e}s-Ancos.
\newblock A fast and accurate deep learning method for strawberry instance segmentation.
\newblock \emph{Computers and Electronics in Agriculture}, 178:\penalty0 105736, 2020.

\bibitem[Plummer et~al.(2015)Plummer, Wang, Cervantes, Caicedo, Hockenmaier, and Lazebnik]{plummer2015flickr30k}
B.~A. Plummer, L.~Wang, C.~M. Cervantes, J.~C. Caicedo, J.~Hockenmaier, and S.~Lazebnik.
\newblock Flickr30k entities: Collecting region-to-phrase correspondences for richer image-to-sentence models.
\newblock In \emph{Proceedings of the IEEE international conference on computer vision}, pages 2641--2649, 2015.

\bibitem[Rai and Sun(2024)]{rai2024weedvision}
N.~Rai and X.~Sun.
\newblock Weedvision: A single-stage deep learning architecture to perform weed detection and segmentation using drone-acquired images.
\newblock \emph{Computers and Electronics in Agriculture}, 219:\penalty0 108792, 2024.

\bibitem[Rezatofighi et~al.(2019)Rezatofighi, Tsoi, Gwak, Sadeghian, Reid, and Savarese]{rezatofighi2019generalized}
H.~Rezatofighi, N.~Tsoi, J.~Gwak, A.~Sadeghian, I.~Reid, and S.~Savarese.
\newblock Generalized intersection over union: A metric and a loss for bounding box regression.
\newblock In \emph{Proceedings of the IEEE/CVF conference on computer vision and pattern recognition}, pages 658--666, 2019.

\bibitem[Sa et~al.(2016)Sa, Ge, Dayoub, Upcroft, Perez, and McCool]{sa2016deepfruits}
I.~Sa, Z.~Ge, F.~Dayoub, B.~Upcroft, T.~Perez, and C.~McCool.
\newblock Deepfruits: A fruit detection system using deep neural networks.
\newblock \emph{sensors}, 16\penalty0 (8):\penalty0 1222, 2016.

\bibitem[Sarig(1993)]{sarig1993robotics}
Y.~Sarig.
\newblock Robotics of fruit harvesting: A state-of-the-art review.
\newblock \emph{Journal of agricultural engineering research}, 54\penalty0 (4):\penalty0 265--280, 1993.

\bibitem[Shao et~al.(2019)Shao, Li, Zhang, Peng, Yu, Zhang, Li, and Sun]{shao2019objects365}
S.~Shao, Z.~Li, T.~Zhang, C.~Peng, G.~Yu, X.~Zhang, J.~Li, and J.~Sun.
\newblock Objects365: A large-scale, high-quality dataset for object detection.
\newblock In \emph{Proceedings of the IEEE/CVF international conference on computer vision}, pages 8430--8439, 2019.

\bibitem[Shi et~al.(2020)Shi, Li, and Yamaguchi]{shi2020attribution}
R.~Shi, T.~Li, and Y.~Yamaguchi.
\newblock An attribution-based pruning method for real-time mango detection with yolo network.
\newblock \emph{Computers and electronics in agriculture}, 169:\penalty0 105214, 2020.

\bibitem[Song et~al.(2023)Song, Wang, Cai, Mondal, and Sahoo]{song2023comprehensive}
Y.~Song, T.~Wang, P.~Cai, S.~K. Mondal, and J.~P. Sahoo.
\newblock A comprehensive survey of few-shot learning: Evolution, applications, challenges, and opportunities.
\newblock \emph{ACM Computing Surveys}, 55\penalty0 (13s):\penalty0 1--40, 2023.

\bibitem[Sun et~al.(2017)Sun, Shrivastava, Singh, and Gupta]{sun2017revisiting}
C.~Sun, A.~Shrivastava, S.~Singh, and A.~Gupta.
\newblock Revisiting unreasonable effectiveness of data in deep learning era.
\newblock In \emph{Proceedings of the IEEE international conference on computer vision}, pages 843--852, 2017.

\bibitem[Syal et~al.(2014)Syal, Garg, and Sharma]{syal2014apple}
A.~Syal, D.~Garg, and S.~Sharma.
\newblock Apple fruit detection and counting using computer vision techniques.
\newblock In \emph{2014 IEEE International Conference on Computational Intelligence and Computing Research}, pages 1--6. IEEE, 2014.

\bibitem[Terven and Cordova-Esparza(2023)]{terven2023comprehensive}
J.~Terven and D.~Cordova-Esparza.
\newblock A comprehensive review of yolo: From yolov1 to yolov8 and beyond.
\newblock \emph{arXiv preprint arXiv:2304.00501}, 2023.

\bibitem[Tian et~al.(2019)Tian, Yang, Wang, Wang, Li, and Liang]{tian2019apple}
Y.~Tian, G.~Yang, Z.~Wang, H.~Wang, E.~Li, and Z.~Liang.
\newblock Apple detection during different growth stages in orchards using the improved yolo-v3 model.
\newblock \emph{Computers and electronics in agriculture}, 157:\penalty0 417--426, 2019.

\bibitem[Tian et~al.(2020)Tian, Shen, Chen, and He]{tian2020fcos}
Z.~Tian, C.~Shen, H.~Chen, and T.~He.
\newblock Fcos: A simple and strong anchor-free object detector.
\newblock \emph{IEEE Transactions on Pattern Analysis and Machine Intelligence}, 44\penalty0 (4):\penalty0 1922--1933, 2020.

\bibitem[Ukwuoma et~al.(2022)Ukwuoma, Zhiguang, Bin~Heyat, Ali, Almaspoor, and Monday]{ukwuoma2022recent}
C.~C. Ukwuoma, Q.~Zhiguang, M.~B. Bin~Heyat, L.~Ali, Z.~Almaspoor, and H.~N. Monday.
\newblock Recent advancements in fruit detection and classification using deep learning techniques.
\newblock \emph{Mathematical Problems in Engineering}, 2022:\penalty0 1--29, 2022.

\bibitem[Wada(2011)]{Wada_Labelme_Image_Polygonal}
K.~Wada.
\newblock {Labelme: Image Polygonal Annotation with Python}, 2011.
\newblock URL \url{https://github.com/wkentaro/labelme}.

\bibitem[Wang et~al.(2024)Wang, Hasler, Tanneberg, Ocker, Joublin, Ceravola, Deigmoeller, and Gienger]{wang2024large}
C.~Wang, S.~Hasler, D.~Tanneberg, F.~Ocker, F.~Joublin, A.~Ceravola, J.~Deigmoeller, and M.~Gienger.
\newblock Large language models for multi-modal human-robot interaction.
\newblock \emph{arXiv preprint arXiv:2401.15174}, 2024.

\bibitem[Wang et~al.(2020)Wang, Yao, Kwok, and Ni]{wang2020generalizing}
Y.~Wang, Q.~Yao, J.~T. Kwok, and L.~M. Ni.
\newblock Generalizing from a few examples: A survey on few-shot learning.
\newblock \emph{ACM computing surveys (csur)}, 53\penalty0 (3):\penalty0 1--34, 2020.

\bibitem[Williams et~al.(2023)Williams, MacFarlane, and Britten]{williams2023leaf}
D.~Williams, F.~MacFarlane, and A.~Britten.
\newblock Leaf only sam: a segment anything pipeline for zero-shot automated leaf segmentation.
\newblock \emph{arXiv preprint arXiv:2305.09418}, 2023.

\bibitem[Wolf et~al.(2019)Wolf, Debut, Sanh, Chaumond, Delangue, Moi, Cistac, Rault, Louf, Funtowicz, et~al.]{wolf2019huggingface}
T.~Wolf, L.~Debut, V.~Sanh, J.~Chaumond, C.~Delangue, A.~Moi, P.~Cistac, T.~Rault, R.~Louf, M.~Funtowicz, et~al.
\newblock Huggingface's transformers: State-of-the-art natural language processing.
\newblock \emph{arXiv preprint arXiv:1910.03771}, 2019.

\bibitem[Xiao et~al.(2023)Xiao, Wang, Xu, and Zhang]{xiao2023fruit}
F.~Xiao, H.~Wang, Y.~Xu, and R.~Zhang.
\newblock Fruit detection and recognition based on deep learning for automatic harvesting: an overview and review.
\newblock \emph{Agronomy}, 13\penalty0 (6):\penalty0 1625, 2023.

\bibitem[Xu et~al.(2022)Xu, Yoon, Fuentes, Yang, and Park]{xu2022style}
M.~Xu, S.~Yoon, A.~Fuentes, J.~Yang, and D.~S. Park.
\newblock Style-consistent image translation: A novel data augmentation paradigm to improve plant disease recognition.
\newblock \emph{Frontiers in Plant Science}, 12:\penalty0 773142, 2022.

\bibitem[Yang et~al.(2023)Yang, Dai, Wu, Bist, Subedi, Sun, Lu, Li, Liu, and Chai]{yang2023sam}
X.~Yang, H.~Dai, Z.~Wu, R.~Bist, S.~Subedi, J.~Sun, G.~Lu, C.~Li, T.~Liu, and L.~Chai.
\newblock Sam for poultry science.
\newblock \emph{arXiv preprint arXiv:2305.10254}, 2023.

\bibitem[Zhang et~al.(2023)Zhang, Han, Qiao, Kim, Bae, Lee, and Hong]{zhang2023faster}
C.~Zhang, D.~Han, Y.~Qiao, J.~U. Kim, S.-H. Bae, S.~Lee, and C.~S. Hong.
\newblock Faster segment anything: Towards lightweight sam for mobile applications.
\newblock \emph{arXiv preprint arXiv:2306.14289}, 2023.

\bibitem[Zhang et~al.(2022)Zhang, Li, Liu, Zhang, Su, Zhu, Ni, and Shum]{zhang2022dino}
H.~Zhang, F.~Li, S.~Liu, L.~Zhang, H.~Su, J.~Zhu, L.~M. Ni, and H.-Y. Shum.
\newblock Dino: Detr with improved denoising anchor boxes for end-to-end object detection.
\newblock \emph{arXiv preprint arXiv:2203.03605}, 2022.

\bibitem[Zhang et~al.(2021)Zhang, Lammers, Chu, Li, and Lu]{zhang2021system}
K.~Zhang, K.~Lammers, P.~Chu, Z.~Li, and R.~Lu.
\newblock System design and control of an apple harvesting robot.
\newblock \emph{Mechatronics}, 79:\penalty0 102644, 2021.

\bibitem[Zhao et~al.(2016)Zhao, Gong, Huang, and Liu]{zhao2016review}
Y.~Zhao, L.~Gong, Y.~Huang, and C.~Liu.
\newblock A review of key techniques of vision-based control for harvesting robot.
\newblock \emph{Computers and Electronics in Agriculture}, 127:\penalty0 311--323, 2016.

\bibitem[Zhou et~al.(2022)Zhou, Wang, Au, Kang, and Chen]{zhou2022intelligent}
H.~Zhou, X.~Wang, W.~Au, H.~Kang, and C.~Chen.
\newblock Intelligent robots for fruit harvesting: Recent developments and future challenges.
\newblock \emph{Precision Agriculture}, 23\penalty0 (5):\penalty0 1856--1907, 2022.

\bibitem[Zhu et~al.(2023)Zhu, Li, Liu, Ma, and Wang]{zhu2023survey}
X.~Zhu, J.~Li, Y.~Liu, C.~Ma, and W.~Wang.
\newblock A survey on model compression for large language models.
\newblock \emph{arXiv preprint arXiv:2308.07633}, 2023.

\bibitem[Zhuang et~al.(2020)Zhuang, Qi, Duan, Xi, Zhu, Zhu, Xiong, and He]{zhuang2020comprehensive}
F.~Zhuang, Z.~Qi, K.~Duan, D.~Xi, Y.~Zhu, H.~Zhu, H.~Xiong, and Q.~He.
\newblock A comprehensive survey on transfer learning.
\newblock \emph{Proceedings of the IEEE}, 109\penalty0 (1):\penalty0 43--76, 2020.

\end{thebibliography}
\end{document}